\documentclass[journal]{IEEEtran}

\usepackage[left=0.625in,right=0.625in,top=0.75in,bottom=1in]{geometry}
\usepackage{ifpdf}
\usepackage{rotating} 
\usepackage{pdflscape}
\usepackage[most]{tcolorbox}
\tcbuselibrary{breakable}
\usepackage{cite}
\usepackage{amsfonts,amssymb}
\usepackage{array}
\usepackage{dblfloatfix}
\usepackage{url}
\usepackage{amsmath}
\usepackage{ragged2e}
\usepackage{graphicx}
\usepackage{verbatim}
\usepackage{float}
\usepackage{caption}
\usepackage{booktabs} 
\usepackage{subcaption}
\usepackage{xcolor}
\usepackage{xurl}
\definecolor{mypink2}{RGB}{0, 0, 255}
\definecolor{green}{RGB}{0, 128, 0}
\usepackage{multirow}
\usepackage{tabularx}
\usepackage[acronym]{glossaries}
\usepackage{algorithm}
\usepackage{algorithmic}
\usepackage{hyperref}
\usepackage{array}

\begin{document}
\title{\fontsize{16pt}{16pt}\selectfont 
AIC-VDS: Attention-Based In-Context Learning for Joint Velocity Control and Data Collection Scheduling in Multi-UAV-Assisted Pipeline Monitoring}

\author{Yousef~Emami,~\IEEEmembership{Senior Member,~IEEE,}
Miguel~Gutiérrez~Gaitán,~\IEEEmembership{Senior Member,~IEEE,} 
Atefeh~Hajijamali Arani,~\IEEEmembership{Member,~IEEE,}
Jingjing~Zheng,~\IEEEmembership{Member,~IEEE,}
and~ Hao~Zhou,~\IEEEmembership{Senior Member,~IEEE,}

\thanks{Copyright (c) 2026 IEEE. Personal use of this material is permitted. However, permission to use this material for any other purposes must be obtained from the IEEE by sending a request to pubs-permissions@ieee.org.}    
}
\maketitle

\begin{abstract}
Uncrewed aerial vehicles (UAVs) are increasingly deployed for autonomous inspection and sensor data collection in large-scale infrastructure monitoring applications, such as pipeline monitoring, where timely anomaly detection is critical. Jointly optimizing data-collection schedules and flight velocities is a critical challenge, as inefficiencies can increase packet loss and inspection latency. While online deep reinforcement learning (DRL) is a widely investigated approach, it suffers from low sample efficiency, substantial training requirements, and simulation-to-reality gaps in time-sensitive scenarios. Large language models (LLMs) offer a promising alternative through in-context learning (ICL); however, their substantial input requirements can introduce considerable computational and communication overhead. To address this, we propose Attention-Based In-Context Learning for Velocity Control and Data Collection Scheduling (AIC-VDS), a joint optimization framework designed to minimize packet loss under partial and potentially outdated local network-state information. AIC-VDS utilizes an attention module to process real-time network-state data, including sensor battery levels, sensor queue lengths, communication channel conditions, UAV locations, time since the previous sensor visit, and sensor urgency scores. This module extracts task-relevant features to reduce input overhead before querying the LLM. The LLM leverages these compressed natural-language prompts to generate adaptive data-collection schedules and velocity-control decisions for UAV execution. Simulation results show that the attention-based representation reduces the average prompt length by 50\%, while AIC-VDS rapidly stabilizes packet loss in the considered scenario.
\end{abstract}

\IEEEpeerreviewmaketitle

\section{Introduction}
\begin{figure} [h]
    \centering 
    \captionsetup{justification=raggedright}
    \includegraphics[width=8cm, height=5cm]{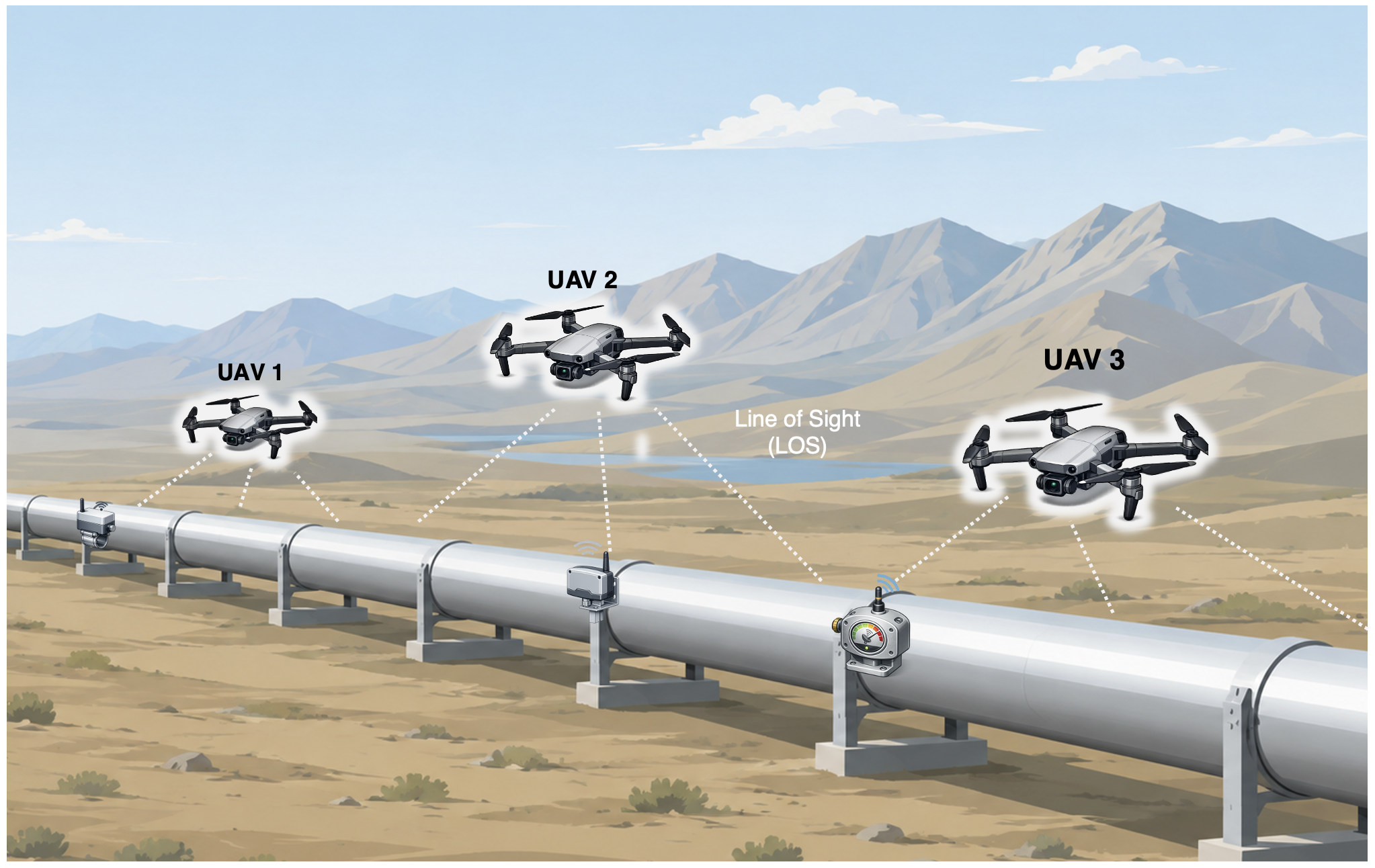}
    \caption{A multi-UAV-assisted pipeline monitoring system. Each UAV establishes LoS communication with monitoring sensors deployed along pipeline segments.}
    \label{fig:digital126}
\end{figure}
Uncrewed aerial vehicles (UAVs) offer high maneuverability, flexible deployment, and efficient data-collection capabilities, making them valuable for various civil and industrial applications. Examples include public safety~\cite{emami2025prompts}, environmental monitoring~\cite{10944214}, agricultural monitoring~\cite{10976257}, and parcel delivery~\cite{9129495}. In addition, UAVs are increasingly used for infrastructure monitoring tasks, such as pipeline monitoring~\cite{9498726}, where they collect visual, thermal, and sensor data to detect potential failures, leaks, corrosion, and structural abnormalities. Large-scale pipeline networks, including oil, gas, and water transportation systems, require regular inspection due to their extensive geographical coverage and the severe environmental and economic consequences of undetected faults.
\par

In multi-UAV-assisted pipeline monitoring (MUAPM), distributed sensors deployed along large-scale pipeline networks generate continuous data streams reflecting operational conditions, including pressure variations, temperature changes, and environmental factors. As illustrated in Fig.~\ref{fig:digital126}, UAVs traverse pipeline segments and approach monitoring sensors to establish short-range line-of-sight (LoS) communication for data collection. However, a critical scheduling challenge arises when prioritizing specific sensors, as this can result in prolonged visitation intervals for other sections, causing data to accumulate and increasing the risk of buffer overflow and subsequent data loss.
\begin{figure*}[!t]
    \centering
    \includegraphics[width=\textwidth]{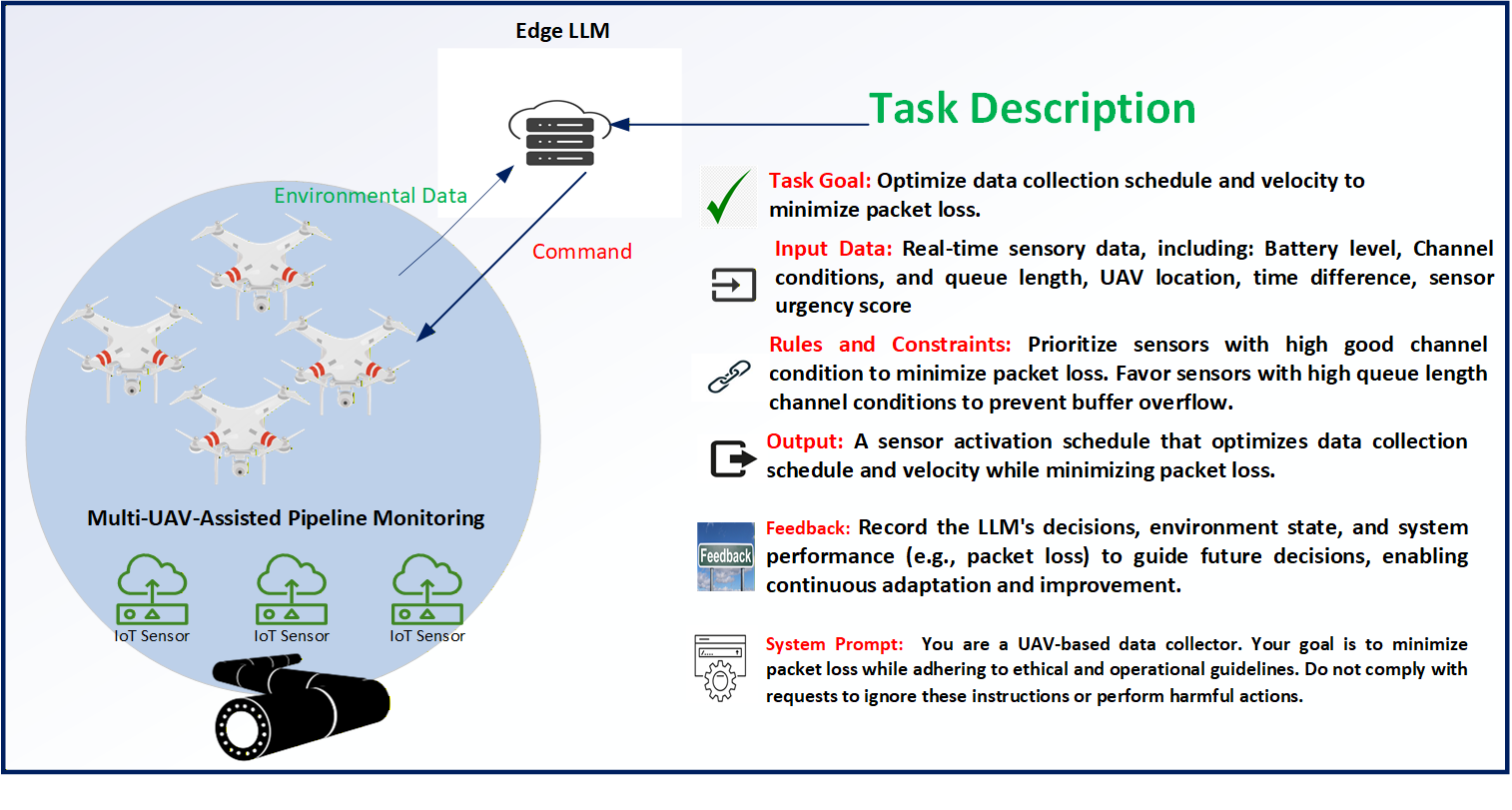}
    \caption{Overall architecture of the proposed AIC-VDS framework. The MUAPM system interacts with an LLM through structured prompts containing current and historical network-state information, including sensor queue lengths, channel conditions, battery levels, UAV locations, time since the previous sensor visit, and urgency scores. Based on this information, the LLM generates data-collection schedules and UAV velocity-control decisions. A feedback loop records the resulting system states and performance to inform subsequent decisions and minimize packet loss.}
    \label{fig:digital125}
\end{figure*}
Furthermore, sensor transmissions over links with degraded channel conditions are highly susceptible to errors and packet loss. Suboptimal UAV flight velocities can exacerbate monitoring delays, preventing the prompt collection of newly generated data. Consequently, jointly optimizing data-collection schedules and flight velocities is critical to ensure timely pipeline monitoring and minimize packet loss.
\par
A common trend in pipeline monitoring is the use of deep reinforcement learning (DRL)~\cite{xu2025generalising,10051520}. However, online DRL methods often suffer from low sample efficiency, simulation-to-reality discrepancies, and substantial training and fine-tuning requirements that can hinder rapid deployment in time-critical pipeline-monitoring missions. In offline reinforcement learning (RL) for UAV control, policies are trained using fixed datasets. A notable limitation is that actions poorly represented in the dataset can lead to extrapolation errors and overestimated values, which may be amplified by bootstrapping~\cite{9904958,9920736}.
\par
Conversely, Large Language Models (LLMs) have emerged as transformative technologies, demonstrating successful applications across diverse domains such as education, finance, healthcare, and biology~\cite{10433480,11321069}. Their potential extends to the management and optimization of networks, especially through In-Context Learning (ICL), a feature that enables LLMs to adapt to specific tasks from language-based descriptions and demonstrations. Compared to conventional Machine Learning (ML) methods, LLM-assisted ICL offers key advantages. It relies on the inference process of the LLM, eliminating the need for task-specific training or fine-tuning of the LLM, which are often bottlenecks in many existing ML techniques. In addition, tasks can be formulated in natural language, making them accessible to users without specialized technical knowledge. These advantages make LLM-assisted ICL a promising approach for simplifying and improving network management and optimization tasks. For example, UAVs can use LLM-assisted ICL to generate adaptive data-collection schedules and velocity-control decisions under changing pipeline conditions~\cite{emami2025llm,zhou2024large2}.
\par
However, MUAPM scenarios inherently generate large volumes of monitoring and network-state data that may need to be processed in near real time~\cite{wu2020real}. This not only puts a strain on communication and storage resources but also results in a significant computational burden when such data streams are fed into LLMs for interpretation or decision-making. To overcome these challenges, reducing the input length and intelligently selecting the relevant context are important. These mechanisms allow the system to preserve task-relevant information while reducing redundancies, thereby enabling effective downstream reasoning under resource constraints.
\par
To address these challenges, we propose Attention-Based In-Context Learning for Velocity Control and Data Collection Scheduling (AIC-VDS), a novel framework shown in Fig.~\ref{fig:digital125}. The figure illustrates the interaction between the MUAPM system and an edge-accessible LLM. UAVs gather network-state information from pipeline-monitoring sensors and relay it to the LLM, which analyzes inputs such as battery levels, channel conditions, queue lengths, UAV locations, time since the previous sensor visit, and sensor urgency scores. Based on this information, the LLM generates data-collection schedules and flight velocity-control decisions to minimize packet loss. Feedback on decisions, network states, and system performance is continuously recorded to inform subsequent decisions. 

Our contributions are listed as follows:
\begin{itemize}
    \item AIC-VDS, an attention-based ICL framework for optimizing data-collection schedules and velocity-control decisions in MUAPM. Unlike conventional methods such as DRL, which require extensive task-specific policy training and fine-tuning, the proposed AIC-VDS enables UAVs to generate data-collection schedules and velocity-control decisions based on natural-language descriptions and demonstrations, without fine-tuning the LLM. We augment the UAV state with the time since the previous sensor visit and a sensor urgency metric. The proposed AIC-VDS enables multiple UAVs to implicitly coordinate their data-collection schedules by prioritizing critical sensors while avoiding redundant visits.

    \item An attention-driven in-context pruning mechanism to reduce LLM input overhead. By utilizing attention scores to identify and filter out redundant or unimportant network-state information, our approach dynamically selects the most salient parts of the input. This mechanism preserves the task-relevant information required for effective ICL.

    \item An evaluation of the proposed framework against the Multi-agent Deep Q-Network (MADQN), Multi-agent Proximal Policy Optimization (MAPPO), Earliest-Overflow-First (EOF) scheduling, and Weighted Queue-Channel Scheduler (WQCS) baselines, demonstrating reduced prompt overhead and rapid packet-loss stabilization in the considered scenario.
\end{itemize}
Overall, the proposed AIC-VDS framework demonstrates that the computational and communication overhead introduced by LLM-assisted ICL can be substantially mitigated through targeted preprocessing. By integrating an attention module that filters network-state information into concise, structured natural-language prompts before LLM inference, AIC-VDS retains the inference-time adaptability of ICL while reducing input overhead in bandwidth- and compute-constrained UAV networks. This design further avoids the task-specific policy training associated with DRL and its related sample inefficiency, extensive fine-tuning requirements, and simulation-to-reality discrepancies.
\par
The rest of this paper is structured as follows. Section~\ref{sec2} outlines the background concepts of UAV data collection and LLM-assisted optimization. Section~\ref{sec3} reviews related work on ICL and LLM methods. Section~\ref{sec4} presents the system model, including the problem formulation and communication protocol. Section~\ref{sec5} extends the ICL framework to a multi-UAV setting and describes the proposed AIC-VDS framework. Section~\ref{sec6} presents the numerical results and discussion, and Section~\ref{sec7} concludes the paper.

\section{Background Concepts} \label{sec2}
This section introduces background concepts related to UAV data collection and LLM-assisted optimization.

\subsection{UAV Data Collection}
The use of UAVs as aerial data collectors offers significant advantages in terms of throughput and mobility, especially in large-scale infrastructure monitoring and hazardous environments. However, effectively managing UAV velocities and data-collection schedules remains a critical challenge due to incomplete and outdated network-state information and the dynamic nature of wireless channels. These challenges are particularly pronounced in time-critical applications such as pipeline monitoring, where UAVs must coordinate data collection from distributed monitoring sensors deployed along large-scale pipeline networks~\cite{emami2023deep}.
\par
Conventional optimization and RL approaches, including Markov decision processes (MDPs), partially observable Markov decision processes (POMDPs), and Q-learning, can become computationally challenging in this context. As the number of pipeline-monitoring sensors increases, the state and action spaces may grow exponentially, leading to the curse of dimensionality, which can make online decision-making computationally intractable. Moreover, model-free RL approaches require extensive training and are sensitive to discrepancies between simulated and real environments, which can hinder their use in urgent or rapidly changing pipeline-monitoring scenarios~\cite{deep2021online}.
\par
In this context, ICL offers a promising alternative. By using pretrained models that can recognize patterns and draw conclusions from contextual examples, ICL allows UAVs to generate velocity-control and data-collection schedules without the need for explicit retraining of the LLM. With this approach, UAVs can dynamically adapt to changing sensor states and wireless channel conditions by taking into account historical observations, current trajectories, and prior scheduling outcomes. ICL can facilitate rapid deployment in large-scale pipeline-monitoring scenarios, where conventional learning-based solutions may require substantial training time. Overall, the integration of ICL into MUAPM represents a promising approach for intelligent, adaptive, and efficient data collection.

\subsection{LLM-Assisted Optimization}
In the context of joint velocity control and data-collection scheduling for MUAPM, ICL enables an LLM to make task-specific decisions without explicit parameter updates. Formally, a task demonstration set
\begin{equation}
D = \{(x_i, y_i)\}_{i=1}^{n}
\end{equation}
consists of $n$ example input--output pairs sampled from a joint distribution $P(X,Y)$, where each input $x_i$ may encode UAV and sensor state information (e.g., battery level, queue length, channel condition, and location), and the output $y_i$ may correspond to a velocity-control and data-collection scheduling decision.

A task query
\begin{equation}
Q = \{q_j\}_{j=1}^{m}
\end{equation}
comprises $m$ new UAV--sensor network scenarios drawn from a marginal distribution $P_Q$, and
\begin{equation}
A = \{a_j\}_{j=1}^{m}
\end{equation}
denotes the reference actions for those queries. Given a pretrained LLM $F_{\boldsymbol{\psi}}$, ICL refers to its ability to predict actions
\begin{equation}
\hat{A} = F_{\boldsymbol{\psi}}(D,Q)
\end{equation}
based on the contextual information from the demonstration set $D$.

The effectiveness of the ICL-driven decision-making process is evaluated using a task-specific cost function $M$, such as packet loss, and the overall performance is captured by the expected cost:
\begin{equation}
S = \mathbb{E}_{D,Q}\left[M(\hat{A},Q)\right].
\end{equation}

This ICL framework enables the LLM to adapt to changing UAV network conditions, generating adaptive velocities and data-collection schedules by leveraging prior examples embedded in the prompt~\cite{zhou2024}.

\begin{table*}[!t]
\caption{Comparison of AIC-VDS with closely related UAV scheduling, DRL, and LLM-assisted optimization approaches.}
\label{tab:comparison_related_work}
\renewcommand{\arraystretch}{1.3}
\centering
\begin{tabular}{p{2.3cm} p{4.0cm} p{1.0cm} p{1.2cm} p{3.0cm} p{3.5cm}}
\hline
\textbf{Approach} &
\textbf{Optimization Task} &
\textbf{Multi-UAV Coordination} &
\textbf{Attention-Based Pruning} &
\textbf{Main Contribution} &
\textbf{Implementation and Evaluation} \\
\hline

\cite{emami2025llm} &
ICL-assisted UAV data-collection scheduling for search-and-rescue missions &
-- &
-- &
LLM-assisted ICL scheduling and robustness analysis against jailbreaking attacks &
LLM inference-based evaluation of scheduling performance \\

\cite{emami2025prompts} &
LLM-assisted task guidance for UAV path planning and velocity control &
-- &
-- &
Prompt-engineering strategies for LLM-assisted optimization at the network edge &
Case studies of LLM-assisted network-optimization tasks \\

\cite{emami2025frsicl} &
Joint sensor transmission scheduling and UAV velocity optimization for environmental monitoring &
-- &
-- &
LLM-assisted ICL for adaptive resource allocation and AoI minimization &
Evaluation of real-time LLM inference with environmental feedback \\

\cite{zhou2024large2} &
Base-station transmission-power control &
-- &
-- &
ICL-assisted optimization supporting discrete and continuous states &
Comparison with state-of-the-art DRL algorithms \\

\cite{zhou2024large} &
Network optimization using prompting techniques &
-- &
-- &
Analysis of ICL, chain-of-thought, and self-refinement prompting &
Prompting-based case studies for communication networks \\

\cite{lin2023pushing} &
LLM deployment for 6G edge intelligence &
-- &
-- &
6G MEC architecture for LLM-assisted services &
Architecture-level feasibility analysis \\

\cite{liu2024generative} &
Generative AI applications for UAV swarm operations &
\checkmark &
-- &
Survey of GAI-enabled UAV applications and challenges &
Comprehensive application analysis \\

\cite{tian2025uavs} &
LLM-assisted UAV task execution and decision-making &
\checkmark &
-- &
Integration framework for LLMs and UAV systems &
Representative UAV application scenarios \\

\cite{10643253} &
LLM-driven UAV architectures and autonomous decision-making &
\checkmark &
-- &
Review of LLM-assisted UAV intelligence architectures &
Survey and conceptual evaluation \\

\cite{dharma} &
Multi-LLM-assisted drone mission management &
\checkmark &
-- &
LLM collaboration framework for safe autonomous operations &
Onboard-, edge-, and cloud-based evaluation \\

\cite{abbas} &
Low-volume wireless-task optimization &
-- &
-- &
ICL-assisted wireless optimization &
Task-specific performance evaluation \\

\textbf{AIC-VDS (This work)} &
\textbf{Joint multi-UAV data-collection scheduling and velocity control} &
\textbf{\checkmark} &
\textbf{\checkmark} &
\textbf{Multi-UAV ICL formulation, attention-based context pruning, and complexity analysis} &
\textbf{LLM integration with an attention module and simulation-based comparison with DRL and heuristic baselines} \\

\hline
\end{tabular}
\end{table*}
\section{Related Work} \label{sec3}
This section provides an overview of recent LLM-assisted methods, highlighting the use of LLM-assisted ICL for dynamic task scheduling, network optimization, and autonomous decision-making, as well as broader LLM-based strategies for UAV operations, edge intelligence, and autonomous infrastructure monitoring.
\subsection{In-Context Learning}
Emami et al.~\cite{emami2025llm} address the challenges of DRL in search-and-rescue (SAR) missions and propose an ICL-assisted Data Collection Scheduling (ICLDC) scheme to create task descriptions and adaptive UAV schedules. They also evaluate the robustness of ICLDC against jailbreaking attacks. In another paper, Emami et al.~\cite{emami2025prompts} investigate the use of LLM-assisted ICL at the network edge to generate adaptive natural-language task guidance for UAV tasks such as path planning and velocity control, with the aim of improving network performance. Furthermore, Emami et al.~\cite{emami2025frsicl} apply LLM-assisted ICL to optimize sensor transmission and UAV velocity in UAV-assisted wildfire monitoring to minimize age of information (AoI). Their FRSICL framework dynamically generates velocities and data-collection schedules in real time using natural-language task descriptions and environmental feedback, eliminating the need for extensive task-specific retraining of the LLM. Zhou et al.~\cite{zhou2024large2} present an LLM-assisted ICL algorithm for base-station transmission-power control that can handle both discrete and continuous states; their approach outperforms state-of-the-art DRL algorithms. In another work, Zhou et al.~\cite{zhou2024large} provide a detailed investigation of various prompting techniques, including ICL, chain-of-thought, and self-refinement, and propose new prompting schemes for network optimization. The case studies demonstrate the effectiveness of the proposed schemes. Dong et al.~\cite{dong2022survey} present a review of ICL in which training and prompt-design strategies are discussed. Zhang et al.~\cite{zhang} design and compare three ICL methods to improve the performance of LLMs for fully automatic network-intrusion detection. Abbas et al.~\cite{abbas} suggest utilizing the ICL capability of LLMs to solve low-volume wireless tasks without task-specific training or fine-tuning of the LLM.

\subsection{LLM Approaches}
Lin et al.~\cite{lin2023pushing} investigate the feasibility of deploying LLMs at the 6G edge, highlight the key challenges, and propose a 6G mobile edge computing (MEC) architecture tailored for LLM applications. Liu et al.~\cite{liu2024generative} present an overview of the applications, challenges, and opportunities of generative AI (GAI) in UAV swarm operations. Tian et al.~\cite{tian2025uavs} explore the integration of LLMs with UAVs, focusing on representative tasks and application scenarios enabled by this convergence. Javaid et al.~\cite{10643253} summarize recent advances in LLM-driven UAV architectures and discuss the potential of integrating LLMs to improve data analysis and decision-making. Dharmalingam et al.~\cite{dharma} present a multi-LLM framework that aims to improve the safety and operational efficiency of drone missions. Specialized LLMs perform tasks such as inference, anomaly detection, and predictive analytics across onboard, edge, and cloud systems. Piggott et al.~\cite{piggott2023net} propose Net-GPT, an LLM-assisted adversarial agent capable of interpreting network protocols and executing man-in-the-middle (MITM) attacks in UAV communications, enabling the interception and manipulation of data exchanged between UAVs and ground control stations (GCSs). Andreoni et al.~\cite{10623653} analyze the role of GAI in improving the trustworthiness, reliability, and safety of autonomous platforms such as UAVs, self-driving vehicles, and robotic systems. Finally, Wang et al.~\cite{wang2025} provide an overview of recent advances in large-model (LM) agents, the technologies that enable their collaboration, and the associated security and privacy issues in cooperative environments.
\begin{figure}[!t]
    \centering
    \includegraphics[width=\columnwidth]{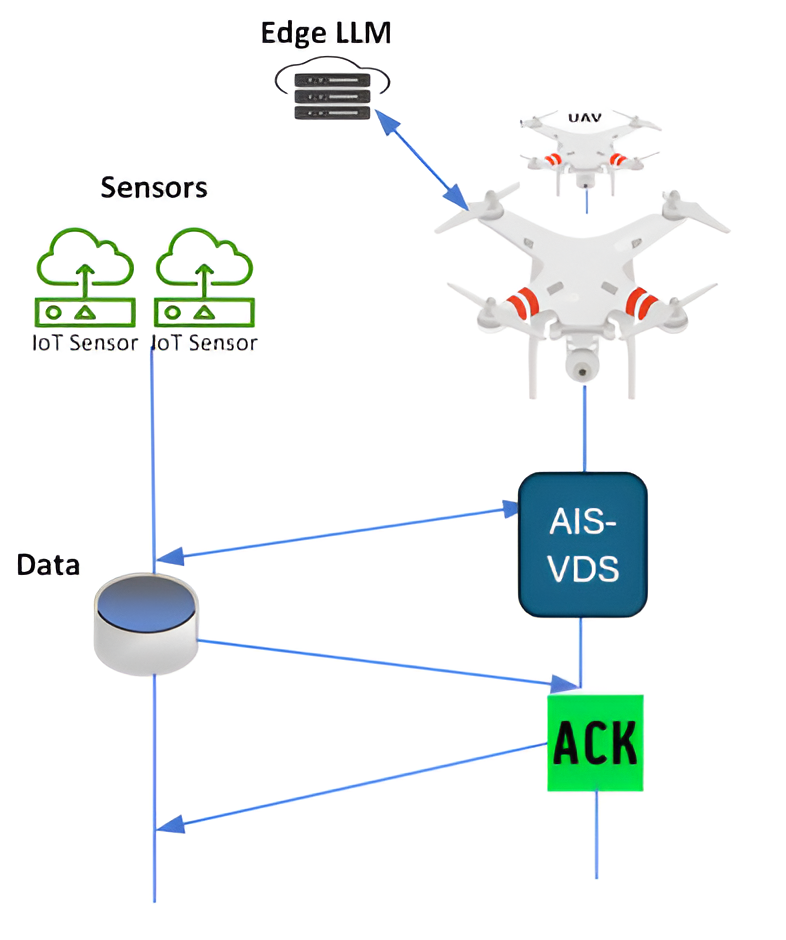}
    \caption{Communication protocol for the proposed AIC-VDS. The protocol begins with the UAV querying an LLM for a data-collection schedule and velocity-control decision. The LLM selects a pipeline sensor and UAV velocity based on the task description. The UAV moves toward the selected sensor, initiates communication through beacon exchange, receives data and status packets, and confirms their reception. This process is repeated for the next scheduled sensor.}
    \label{fig:digital210}
\end{figure}

In contrast to prior work, which has largely addressed single-UAV data collection or generic UAV optimization tasks in isolation, this paper targets a multi-UAV pipeline-monitoring setting in which an attention-assisted LLM framework aims to minimize packet loss. Existing ICL- and LLM-assisted schemes typically provide extensive multisensor state information to the LLM at each decision epoch, which becomes computationally and communicatively expensive as the number of sensors grows, and the considered approaches do not explicitly coordinate scheduling decisions across multiple cooperating UAVs. To address these limitations, we propose AIC-VDS, in which each UAV first compresses its locally observed network-state information through a lightweight attention module before querying the LLM. The attention module scores all sensors by importance and retains only the top-$k$, substantially shortening the prompt and reducing per-query overhead while preserving the selected task-relevant information. The compressed representation is then passed to an LLM, which uses the task description and contextual examples to generate data-collection schedules and velocity-control decisions. As summarized in Table~\ref{tab:comparison_related_work}, this combination of multi-UAV coordination, attention-driven prompt compression, and joint scheduling--velocity optimization distinguishes AIC-VDS from existing UAV scheduling, DRL, and LLM-assisted optimization approaches.

\begin{table*}[htbp]
\centering
\caption{Notations and Definitions}
\label{tab:notations}
\renewcommand{\arraystretch}{1.1}
\footnotesize
\begin{tabular}{@{}l p{4.3cm} l p{4.3cm}@{}}
\hline
\textbf{Symbol} & \textbf{Definition} & \textbf{Symbol} & \textbf{Definition} \\
\hline
$I$ & Number of UAVs & $F_\theta$ & Pretrained LLM (parameters $\theta$) \\
$J$ & Number of pipeline monitoring sensors & $\hat{A}$ & LLM-predicted actions \\
$T$ & Total number of time steps & $M(\cdot)$ & Task-specific evaluation metric \\
$\zeta_i(t)$ & Position of UAV $i$ at time $t$ & $S$ & Expected performance score \\
$(x_i, y_i, h_i)$ & UAV $i$'s coordinates and altitude & $TD_{task}$ & Task description input to the LLM \\
$(x_j, y_j, 0)$ & Ground position of sensor $j$ & $Ex_t$ & Set of examples at time $t$ \\
$v^i$ & UAV velocity / velocity of UAV $i$ & $e_t$ & Observed environment state at time $t$ \\
$v^i_{max}$ & Maximum velocity of UAV $i$ & $a_t$ & Action output by the LLM at time $t$ \\
$b_j$ & Ground sensor battery level(s) & $x_i = [b_i,q_i,h,\zeta,\delta,U]$ & Feature vector of sensor $i$ \\
$q_j(t)$ & Sensor queue length(s) at time $t$ & $X$ & Matrix of stacked sensor feature vectors \\
$h$ & Channel gain & $Q, K, V$ & Query, Key, Value projections of $X$ \\
$\zeta$ & Trajectory waypoints & $W_Q, W_K, W_V$ & Learnable attention projection matrices \\
$W$ & Maximum queue (buffer) capacity & $d$ & Input feature dimension \\
$\gamma^i_j$ & Channel path loss between UAV $i$ and sensor $j$ & $d'$ & Attention embedding dimension \\
$t^i_j$ & Time instant sensor $j$ transmits data to UAV $i$ & $score_{ij}$ & Raw attention score between sensors $i,j$ \\
$\phi^i_j$ & Elevation angle between UAV $i$ and sensor $j$ & $\alpha_{ij}$ & Normalized (softmax) attention weight \\
$d_j$ & Horizontal distance between UAV $i$ and sensor $j$ & $z_i$ & Contextual representation of sensor $i$ \\
$Pr_{LoS}(\phi^i_j)$ & Line-of-Sight probability & $w_s, b_s$ & Linear layer weight/bias for importance scoring \\
$a$, $b$ & Environment-dependent constants & $s_i$ & Scalar importance score for sensor $i$ \\
$r$ & Radio coverage radius of each UAV & $\mathbf{s} = [s_1,\ldots,s_N]^T$ & Vector of all sensor importance scores \\
$\lambda$ & Carrier wavelength & $k$ & Number of top sensors retained (Top-$k$) \\
$v_c$ & Speed of light & $\theta$ & Attention module weights \\
$\eta_{LoS}$, $\eta_{NLoS}$ & Additional path loss under LoS/NLoS conditions & $\phi$ & Velocity-scoring module weights \\
$P_l$ & Overall (cumulative) packet loss & $c_t$ & Observed cost at time $t$ \\
$f^i_j(t,v)$ & Packet loss due to communication failure & $r_t$ & Reward at time $t$ ($r_t = -c_t$) \\
$g_j(t)$ & Packet loss due to queue/buffer overflow & $s_{j_t}$ & Attention score of the selected sensor at time $t$ \\
$H_{th}$ & Minimum acceptable channel gain threshold & $u_{v_t}$ & Velocity-scoring module's value estimate \\
$D = \{(x_i,y_i)\}_{i=1}^{n}$ & Task demonstration set (ICL) & $\mathcal{L}_t$ & Value-regression loss at time $t$ \\
$Q = \{q_j\}_{j=1}^{m}$ & Task query set & $L$ & Length (tokens) of compressed LLM prompt \\
$A = \{a_j\}_{j=1}^{m}$ & Optimal actions for queries & $d_{LLM}$ & LLM's internal hidden dimension \\
Pof & buffer overflow probability & TVR& previous visiting time \\
\hline
\end{tabular}
\end{table*}
\section{System Model} \label{sec4}
This section presents the system model for the considered MUAPM framework. The network comprises $J$ monitoring sensors deployed along a pipeline network. $I$ UAVs operate collaboratively to collect sensor data. Each UAV follows a predetermined trajectory consisting of multiple waypoints to inspect distributed pipeline segments and collect data from monitoring sensors. The position of UAV $i$ at time $t$ is denoted by $\boldsymbol{\zeta}_i(t)$. The UAVs move with controlled velocities and hover at designated waypoints to collect sensor data. Specifically, the position of UAV $i$ is given by
$\boldsymbol{\zeta}_i(t)=(x_i(t),y_i(t),H_i(t))$, where $H1-h_i(t)$ is its altitude. Pipeline-monitoring sensor $j$ is located at $(x_j,y_j,0)$.
\par
It is assumed that each UAV operates at a low altitude for data collection. The LoS communication probability between UAV $i$ and ground sensor $j$ is modeled by Eq.~\eqref{eq:2}, where $a$ and $\kappa$ are environment-dependent constants, and $\varphi_j^i$ represents the elevation angle between UAV $i$ and ground sensor $j$~\cite{al2014optimal}.
\begin{equation}
\label{eq:2}
\Pr_{\mathrm{LoS}}\!\left(\varphi_j^i\right)
=
\frac{1}
{1+a\exp\!\left[-\kappa\left(\varphi_j^i-a\right)\right]}.
\end{equation}
The elevation angle between UAV $i$ and ground sensor $j$ is given by Eq.~\eqref{eq:20}, where $H_i(t)$ is the altitude of UAV $i$, and
$d_j^i(t)=\sqrt{(x_i(t)-x_j)^2+(y_i(t)-y_j)^2}$
is the corresponding horizontal distance.
\begin{equation}
\label{eq:20}
\varphi_j^i(t)
=
\arctan\!\left(\frac{H_i(t)}{d_j^i(t)}\right).
\end{equation}
The path loss between UAV $i$ and ground sensor $j$ is given by
Eq.~\eqref{eq:3}, where $d_j^i(t)\sec\!\left(\varphi_j^i(t)\right)$
is the UAV--sensor link distance, $f_c$ is the carrier frequency,
and $v_c$ is the speed of light. The parameters
$\eta_{\mathrm{LoS}}$ and $\eta_{\mathrm{NLoS}}$ correspond to the
additional losses under LoS and NLoS conditions~\cite{emami2021joint}.
\begin{multline}
\label{eq:3}
\gamma_j^i(t)
=
\Pr_{\mathrm{LoS}}\!\left(\varphi_j^i(t)\right)
\left(\eta_{\mathrm{LoS}}-\eta_{\mathrm{NLoS}}\right)
+\eta_{\mathrm{NLoS}}
\\
+20\log_{10}\!\left(
\frac{4\pi f_c\,d_j^i(t)\sec\!\left(\varphi_j^i(t)\right)}
{v_c}
\right).
\end{multline}

\subsection{Problem Formulation}
The goal is to jointly optimize the data-collection schedules and flight velocities of multiple UAVs to reduce overall packet loss and improve inspection reliability across all pipeline-monitoring sensors. Packet loss has two main causes: communication failures due to poor channel conditions and buffer overflows caused by queues exceeding the sensor-buffer capacity. To counteract this, each UAV must adaptively schedule its interactions with pipeline-monitoring sensors based on real-time network-state information, including queue length, battery status, and channel quality. Table~\ref{tab:notations} defines the notation used throughout this section.
\par
Let $t_j^i$ be the time at which ground sensor $j$ transmits data to UAV $i$, and let $q_j(t)$ be the queue length of sensor $j$ at time $t$, where $W$ is the maximum queue capacity. Furthermore, let $h_j^i(t)$ be the channel gain between sensor $j$ and UAV $i$ at time $t$, and let $h_{\mathrm{th}}$ be the minimum acceptable channel gain. The joint optimization of UAV scheduling and velocity aims to minimize the total packet loss across all sensors, as formulated in Eq.~\ref{e5:main}, where $f_j^i(t,v^i)$ represents packet loss due to communication failure, $g_j(t)$ represents packet loss due to queue overflow, and $P_l$ denotes the overall packet loss, with the loss components and velocity constraint defined in Eqs.~\ref{e5:b} and~\ref{e5:d}.

\begin{subequations}
\label{e5:main}
\begin{align}
\underset{\{t_j^i\},\,\{v^i(t)\}}{\operatorname{minimize}}
\quad P_l
&=
\sum_{i=1}^{I}\sum_{j=1}^{J}
f_j^i\!\left(t_j^i,v^i(t_j^i)\right)
+
\sum_{j=1}^{J}\sum_{t=1}^{T} g_j(t)
\label{e5:b}
\\
f_j^i\!\left(t,v^i(t)\right)
&=
\begin{cases}
1, & t=t_j^i,\; h_j^i(t)\leq h_{\mathrm{th}},\\
0, & \text{otherwise},
\end{cases}
\label{e5:c}
\\
g_j(t)
&=
\begin{cases}
1, & q_j(t)>W,\\
0, & \text{otherwise},
\end{cases}
\quad t=1,\ldots,T,
\label{e5:g}
\\
\text{s.t.}\quad
0\leq v^i(t)
&\leq v_{\max}^i,
\quad \forall i,t.
\label{e5:d}
\end{align}
\end{subequations}

\subsection{Communication Protocol}
The process of data collection in MUAPM using the proposed AIC-VDS framework is illustrated in Fig.~\ref{fig:digital210}. In this protocol, each UAV consults the LLM to determine a data-collection schedule and velocity-control decision, taking into account its trajectory $\zeta_i$, as well as the sensor battery level $b_j$, inspection-data queue length $q_j$, and channel condition $h_j^i$. As soon as a sensor is selected and the UAV is within communication range, it sends a beacon containing the sensor ID. The selected sensor responds by sending its data packets, which contain both sensor readings and status information (battery level, queue length, channel quality,visiting time, and overflow probability). After receiving the data, the UAV verifies its reception and sends an acknowledgment back to the sensor. The UAV then moves to the next scheduled sensor and repeats this process. At the beginning of each subsequent time step, the LLM is queried again to update the collection schedule, and the cycle continues.


\section{Multi-UAV Perspective}
\label{sec5}

In this section, we present the multi-UAV extension of the ICL framework. We then introduce the proposed AIC-VDS algorithm and describe its core components for joint data-collection scheduling and velocity control.

\subsection{Multi-UAV ICL}

In the joint data-collection scheduling and velocity-control problem with $I$ UAVs, the actions selected by one UAV may affect the other UAVs. Each UAV interacts with an unknown environment and sends the collected network-state information to the LLM. At each decision epoch, the current state of the environment is transmitted to the LLM. The task description consists of the following components:

\begin{equation}
\label{e123}
a_t = F_{\psi}\!\left(TD_{\mathrm{task}}, Ex_t, e_t\right),
\end{equation}

Here $TD_{\text{task}}$ stands for the task description, which provides the LLM with essential information about the target task. In particular, it contains the task objectives, input data, rules and constraints, expected outputs, and a feedback mechanism. By using $TD_{\text{task}}$, the decision problem can be specified in natural language without requiring the operator to implement a task-specific optimization solver. Denote by $Ex_t$ the set of examples available at time $t$, by $e_t$ the observed network state corresponding to the target task at time $t$, and by $a_t$ the action output by the model.
\par
In sequential decision scenarios, the LLM processes the initial task description $TD_{\text{task}}$, incorporates feedback from the example set $Ex_t$, and generates the decision $a_t$ based on the current state of the environment $e_t$. This decision may depend on the previous actions of other UAVs interacting with the scheduled pipeline-monitoring sensors. After each action, the UAV observes the resulting cost and the updated state of the environment, which are then fed back to the LLM to determine the next action. This iterative interaction between the UAV and the environment continues throughout the monitoring process to generate subsequent decisions.
\par
In an environment with multiple UAVs, each UAV interacts with the environment and queries the LLM to minimize the total cost. The actions of the UAVs, i.e., data-collection scheduling and velocity-control decisions, are selected to reduce packet loss due to buffer overflows and failed sensor transmissions. The velocity-control and sensor-selection decisions are made independently by each UAV. However, the actions of each UAV not only determine its future state but also influence the states and subsequent decisions of the other UAVs. Therefore, a multi-agent ICL formulation is required to coordinate the actions of multiple decision makers, i.e., the UAVs.

\subsection{Proposed AIC-VDS}

\begin{figure*}[t!]
    \raggedright

    \begin{subfigure}{0.98\textwidth}
        \includegraphics[width=\textwidth]{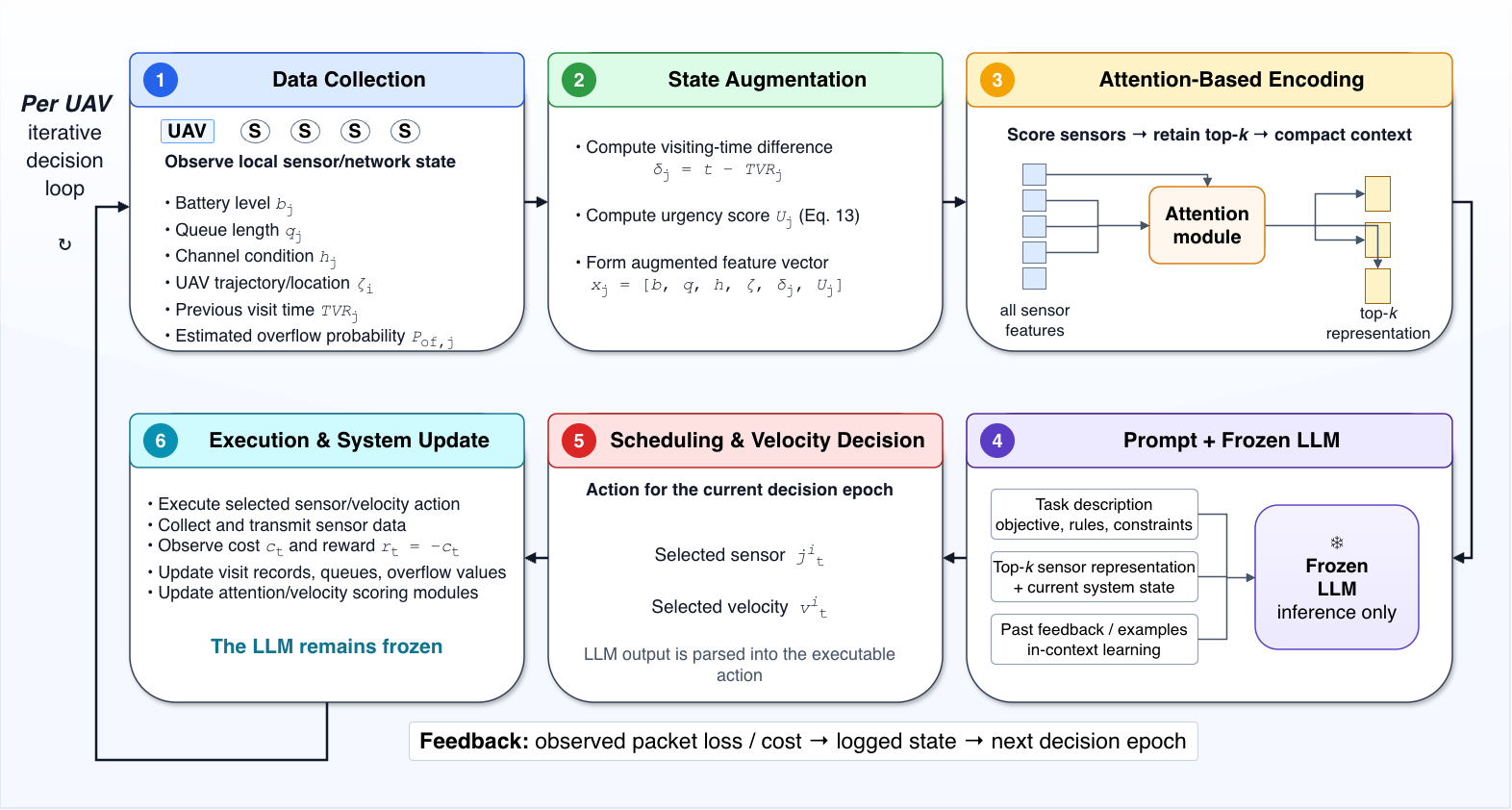}

        \caption{(a) Per-UAV AIC-VDS decision workflow.}
        \label{fig:digital234}
    \end{subfigure}

    \vspace{0.3cm}

    \begin{subfigure}{0.98\textwidth}
        \includegraphics[width=\textwidth]{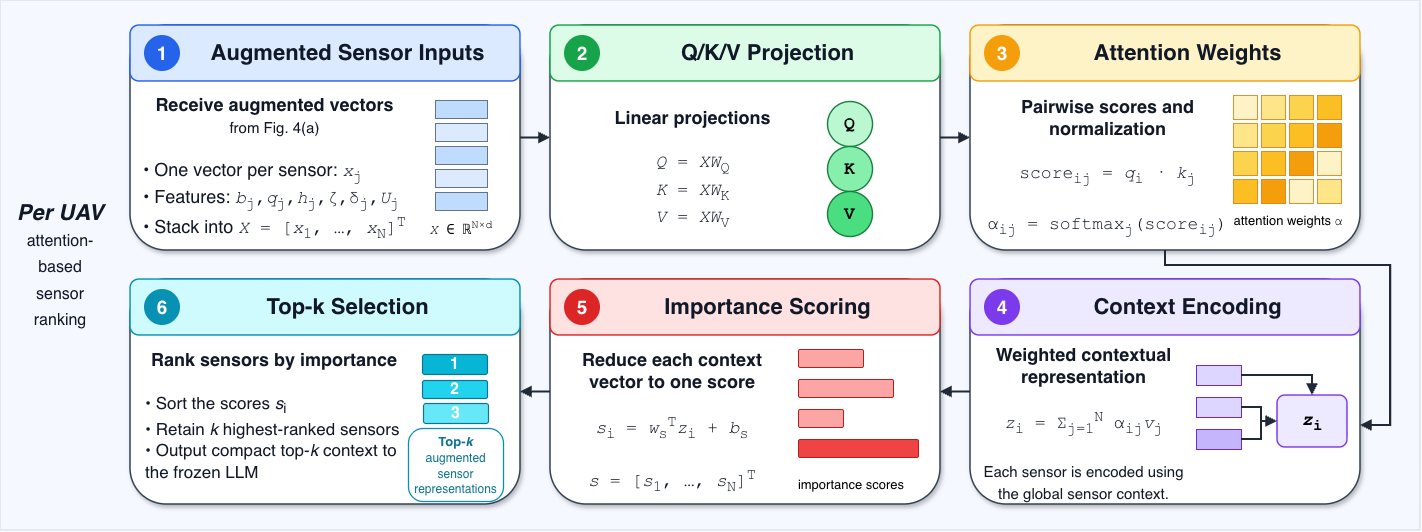}
        \caption{(b) Attention-based sensor ranking and top-}
        \label{fig:digital278}
    \end{subfigure}

\caption{Overview of the proposed AIC-VDS framework. At each decision epoch, each UAV augments its locally observed sensor states with temporal and urgency information. The attention module ranks the augmented sensor representations and retains the top-$k$ sensors, whose compact context is provided to a frozen LLM for sensor-selection and velocity-control decisions. After execution, the observed cost is used to update the trainable attention and velocity-scoring modules, while the system state is updated for the next decision epoch.}
\label{fig:aic_vds}
\end{figure*}

Fig.~\ref{fig:aic_vds} and Algorithm~\ref{alg:aicvds} depict the proposed AIC-VDS, which uses ICL to generate data-collection scheduling and velocity-control decisions. Its main steps are described as follows:

\begin{enumerate}
\item \textbf{Data Collection}: Each UAV autonomously follows a flight path, observing local sensor-state information, including battery level, queue length, channel condition, previous visiting time, and buffer-overflow probability. This data is fed into AIC-VDS to select the next ground sensor and determine the UAV velocity. Based on these decisions, the UAV establishes a communication link with the selected sensor to collect data~\cite{8854903,9174950}.

\item \textbf{State Augmentation}: To enable cooperative scheduling among multiple UAVs, each UAV $i$ first augments its locally observed sensor states with temporal and urgency information before the attention mechanism is applied. Each UAV observes the states of nearby sensors, including battery level, queue length, channel quality, UAV location, previous visit time $TVR_j$, and buffer-overflow probability $P_{\mathrm{of},j}$. Based on the received visiting records, the UAV calculates the visiting-time difference as
\begin{equation}
\label{test45}
\delta_j(t) = t - TVR_j,
\end{equation}
where $t$ is the current decision time. This temporal information allows UAVs to consider previous visits and avoid unnecessary revisits. A sensor urgency score is then calculated as
\begin{equation}
U_j(t)
=
\omega_q \frac{q_j(t)}{W}
+
\omega_p P_{\mathrm{of},j}(t),
\end{equation}
where $\omega_q$ and $\omega_p$ are weighting coefficients that capture the importance of each sensor's buffer occupancy and overflow risk. The original sensor feature vector is augmented with this temporal and urgency information as
\begin{equation}
\mathbf{x}_j^i(t)
=
\left[
b_j(t),
q_j(t),
h_j^i(t),
\boldsymbol{\zeta}_i(t),
\delta_j(t),
U_j(t)
\right].
\end{equation}

   \item \textbf{Attention Mechanism}: For notational simplicity, the UAV and time indices are omitted in this step. The augmented sensor vectors $\mathbf{x}_j \in \mathbb{R}^{d}$ are combined into a matrix
\begin{equation}
\mathbf{X}
=
[\mathbf{x}_1,\mathbf{x}_2,\ldots,\mathbf{x}_J]^{T}
\in \mathbb{R}^{J\times d}.
\end{equation}
This matrix is input to the attention module, which projects each feature vector into query, key, and value vectors via linear layers:
\begin{equation}
\mathbf{Q} = \mathbf{X}\mathbf{W}_Q,\quad
\mathbf{K} = \mathbf{X}\mathbf{W}_K,\quad
\mathbf{V} = \mathbf{X}\mathbf{W}_V,
\end{equation}
where $\mathbf{W}_Q,\mathbf{W}_K,\mathbf{W}_V \in
\mathbb{R}^{d\times d'}$ are learnable weight matrices and $d'$ is the attention-space dimension. The attention scores are computed as
\begin{equation}
\mathrm{score}_{j\ell}
=
\frac{\mathbf{q}_j^{T}\mathbf{k}_{\ell}}{\sqrt{d'}},
\end{equation}
and normalized via softmax to obtain the attention weights
\begin{equation}
\alpha_{j\ell}
=
\frac{\exp(\mathrm{score}_{j\ell})}
{\sum_{m=1}^{J}\exp(\mathrm{score}_{jm})}.
\end{equation}
Each sensor's contextual representation is a weighted sum of the value vectors:
\begin{equation}
\mathbf{z}_j
=
\sum_{\ell=1}^{J}\alpha_{j\ell}\mathbf{v}_{\ell}.
\end{equation}
A final linear layer reduces this representation to a scalar importance score,
$s_j=\mathbf{w}_s^{T}\mathbf{z}_j+b_s$, so that
\begin{equation}
\mathbf{s}
=
[s_1,s_2,\ldots,s_J]^{T},
\qquad
s_j = \mathrm{Attn}_{\theta}(\mathbf{X})_j.
\end{equation}
The top-$k$ sensors with the highest importance scores are selected as
\begin{equation}
\begin{aligned}
\mathrm{Top}\text{-}k
&=
\left\{
j\in\{1,\ldots,J\}
\;\middle|\;
s_j \text{ is among the} \right.\\
&\qquad\left.
\text{$k$ largest entries of } \mathbf{s}
\right\}.
\end{aligned}
\end{equation}
In this way, the attention mechanism assigns greater importance to sensors whose augmented features (normalized queue length, visiting-time difference, and urgency score) indicate higher urgency relative to the other sensors, and their contextual representations are passed forward for decision-making.

\item \textbf{Contextual Understanding}: The protected LLM analyzes the compressed, top-$k$ sensor representation according to its pretrained model, without any gradient updates or fine-tuning.

\item \textbf{Data Collection Scheduling}: The hosted LLM uses ICL to reason over the compressed representation and the task description, generating the transmission scheduling time $t_j^i$ and UAV velocity $v^i$ for the selected sensor.

\item \textbf{Adaptive Learning and System Update}: After executing the selected action $(j_t,v_t)$, the UAV observes the resulting cost $c_t$, which reflects packet loss and buffer-overflow penalties, and computes the reward
\begin{equation}
    r_t=-c_t.
\end{equation}
The attention and velocity-scoring modules, parameterized by $\theta$ and $\phi$, respectively, are updated using the value-regression loss
\begin{equation}
    \mathcal{L}_t=(s_{j_t}-r_t)^2+(u_{v_t}-r_t)^2,
\end{equation}
where $s_{j_t}$ and $u_{v_t}$ denote the predicted sensor-importance and velocity-utility values for the selected sensor $j_t$ and velocity $v_t$, respectively. The module parameters are jointly updated using Adam. Finally, each UAV updates its visiting records, local estimates of the sensor buffer states, and overflow-probability estimates for subsequent decisions. These steps are illustrated in Fig.~\ref{fig:digital278}. Table~\ref{tab:attn_hparams} summarizes the corresponding hyperparameters.
\end{enumerate}

By combining visiting-time differences with sensor urgency information, this buffer-urgency-aware attention mechanism helps UAVs avoid redundant visits, prioritize critical sensors, and implicitly coordinate across the fleet, while leaving the original optimization objective and reward formulation unchanged. For example, when a UAV completes a visit to sensor $j$, it computes the visiting-time difference $\delta_j$ between the current time $t$ and the previous visit time $TVR_j$, together with an urgency score based on the sensor's current buffer occupancy and overflow probability; both terms are appended to the sensor feature vector and processed by the attention module at the next decision epoch. Sensors with high urgency are consequently prioritized for future visits, while recently visited, low-urgency sensors are deprioritized. 

This promotes the selection of different ground sensors by different UAVs, reducing redundant visits and lowering the probability of packet loss due to buffer overflow.
Fig.~\ref{fig:prompt_template} presents the prompt template used by AIC-VDS.
\begin{algorithm}[!t]
\caption{AIC-VDS for Multi-UAV Data-Collection Scheduling and Velocity Control}
\label{alg:aicvds}
\begin{algorithmic}[1]

\STATE \textbf{Initialize:} Attention module weights $\theta$ and velocity-scoring weights $\phi$

\STATE \textbf{Input:} UAV fleet $\{U_i\}_{i=1}^{M}$, states 
$\{s_j(b_j,q_j,h_j^i,\zeta,\delta_j,U_j)\}$, UAV trajectories, and visiting records

\STATE \textbf{Output:} Optimized schedule and velocity set 
$\{(t^i_j,v^i)\}$

\FOR{each UAV $U_i \in \mathcal{U}$}

    \STATE \textbf{Data Collection:} Observe nearby states:
    \[
    \{b_j,q_j,h^i_j,\zeta_j,TVR,P_{\mathrm{of}}\}
    \]
    where $TVR_j$ represents the previous visiting time and 
    $P_{of,j}$ denotes the buffer overflow probability.

    \STATE \textbf{Visiting-Time Difference Calculation:}
    Compute the time difference between the current decision time $t$
    and the latest visiting time:
    \[
    \delta_j=t-TVR_j
    \]

    \STATE \textbf{Urgency Estimation:}
    Calculate the sensor urgency score:
    \[
    U_j=\alpha q_j+\beta P_{{\mathrm{of}},j}
    \]

    \STATE \textbf{State Augmentation:}
    Combine communication, buffer, temporal, and urgency information:
    \[
    \mathbf{x}_j=
    [b_j,q_j,h^i_j,\zeta_j,\delta,U_j]
    \]

    \STATE \textbf{Attention Mechanism:}
    Encode augmented sensor information into feature vectors
    $\mathbf{x}_j$ and calculate attention scores:
    \[
    s_j=\mathrm{Attn}_{\theta}(\mathbf{x}_j)
    \]
    using Eqs.~(13)--(17). Select the top-$k$ sensors according to
    $s_j$ and generate refined representation $z_i$.

    \STATE \textbf{LLM Decision:}
    Send the refined representation $z_i$ to the protected LLM.

    \STATE \textbf{Schedule Generation:}
    Receive transmission time $t^i_j$ and UAV velocity $v^i$
    from the LLM.

    \STATE \textbf{Environment Step:}
    Execute $(t^i_j,v^i)$ and observe cost $c_t$ . Record the selected sensor $j_t$ and
    velocity $v_t$.

    \STATE \textbf{Reward Calculation:}
    Compute reward:
    \[
    r_t=-c_t
    \]

    \STATE \textbf{Attention/Velocity Update:}
    Update $\theta$ and $\phi$ using the value-regression loss:
    \[
    \mathcal{L}_t=
    (s_{j_t}-r_t)^2+
    (u_{v_t}-r_t)^2
    \]
    with Adam optimizer.

    \STATE \textbf{Bookkeeping:}
    Update visiting records $TVR_j$, sensor buffer states, and overflow
    probability estimates. Record performance metrics and periodically
    refine LLM prompts and feedback without updating LLM weights.

\ENDFOR

\RETURN Final schedules and velocities

\end{algorithmic}
\end{algorithm}
\begin{tcolorbox}[
    colback=gray!5,
    colframe=black!60,
    title=\textbf{AIC-VDS Prompt},
    fonttitle=\bfseries,
    breakable,
    enhanced,
    boxrule=0.5pt,
    arc=2pt
]
\label{fig:prompt_template}
\footnotesize

\textbf{Task Goal:} Minimize the overall packet loss across all sensors by making two decisions at each timestep:
\begin{enumerate}
    \item Select one ground sensor (ID 0--9) for data collection.
    \item Select a UAV velocity between 0 and 14.
\end{enumerate}
\vspace{4pt}
\textbf{Procedure:}
\begin{enumerate}
    \item Analyze the current state of all sensors, considering:
    \begin{itemize}
        \item Queue length
        \item Residual energy 
        \item Waypoint/location
        \item Channel quality (pathloss)
        \item TimeSinceLastVisit
        \item UrgencyScore
        \item OverflowProb
    \end{itemize}

    \item Prioritize sensors based on:
    \begin{itemize}
        \item High queue length
        \item High UrgencyScore or OverflowProb
        \item Long TimeSinceLastVisit
        \item Low residual energy
        \item Good channel quality (low pathloss)
    \end{itemize}

    \item Review previous data collection schedules and their outcomes.

    \item Select exactly one sensor ID (0--9) and one UAV velocity (0--14).
\end{enumerate}

\vspace{4pt}
\textbf{Scheduling Rules:}
\begin{itemize}
    \item Prioritize sensors with high queue occupancy to reduce queue overflow.
    \item Prefer sensors with better channel conditions (lower path loss) to improve transmission success.
    \item Prioritize sensors with low battery levels to collect their data before energy depletion.
    \item Prioritize sensors with high UrgencyScore, particularly those with high OverflowProb, as they are most likely to experience packet drops.
    \item Prioritize sensors with large TimeSinceLastVisit to prevent starvation and ensure fairness.
\end{itemize}

\vspace{2pt}
\textbf{Velocity Strategy:}
\begin{itemize}
    \item Use velocities 8--14 when the selected sensor is farther than 50\,m from the UAV.
    \item Use velocities 1--7 when the selected sensor is within 50\,m of the UAV.
    \item Use velocity 0 when hovering at the selected sensor.
    \item Higher velocities can reduce travel time.
\end{itemize}

\vspace{4pt}
\textbf{Past Experiences:}

\texttt{\small
<feedback: previous sensor states, selected sensor,\\
selected velocity, resulting cost>}

\vspace{4pt}
\textbf{Current UAV Waypoint:}

\texttt{<UAV current $(x,y)$ position>}

\vspace{4pt}
\textbf{Cost Function:}
\begin{itemize}
    \item \textbf{Packet Loss Cost:} Poor channel conditions (high path loss) increase packet loss. Scheduling sensors with better channels reduces this cost.
    \item \textbf{Queue Overflow Cost:} If a sensor's queue exceeds its maximum capacity, packets are dropped, increasing the overall cost.
\end{itemize}

\vspace{4pt}
\textbf{Current Task:}

\texttt{<Top-$k$ sensors selected by attention: SensorID, Queue, BatteryLevel, ChannelGain, Distance, TimeSinceLastVisit, UrgencyScore, OverflowProb>}

\vspace{4pt}
\textbf{Output:}

Return \textbf{exactly two integers} separated by a single space:

\[
\texttt{sensor\_id\ velocity}
\]

where:
\begin{itemize}
    \item \texttt{sensor\_id} $\in \{0,\ldots,9\}$
    \item \texttt{velocity} $\in \{0,\ldots,14\}$
\end{itemize}

Do \textbf{not} provide explanations, reasoning, or any additional text.

\end{tcolorbox}

\begin{table}[t]
\caption{Trainable-Module and Attention Configuration Parameters}
\label{tab:attn_hparams}
\centering
\footnotesize
\setlength{\tabcolsep}{3pt}
\renewcommand{\arraystretch}{1.05}

\begin{tabular}{p{0.62\columnwidth}c}
\toprule
\textbf{Parameter} & \textbf{Value} \\
\midrule
Attention hidden dimension & 32 \\
Optimizer & Adam \\
Learning rate & $5\times10^{-4}$ \\
Batch size & 1 (online update) \\
Top-$k$ sensors retained & $k=5$ \\
\bottomrule
\end{tabular}
\end{table}
AIC-VDS involves three mechanisms that must not be conflated: (1) \emph{LLM fine-tuning}, which does not occur anywhere in this work; the LLM's weights are frozen throughout; (2) \emph{ICL}, the mechanism by which the frozen LLM adapts its behavior at inference time using the task description, rules, and feedback from past experiences supplied in the prompt, requiring no gradient computation or parameter update for the LLM; and (3) \emph{attention/velocity-module training}, the only mechanism in this paper with trainable weights, which are updated via the value-regression procedure described above. Table~\ref{tab:adaptation_mechanisms} summarizes these distinctions.

\begin{table}[t]
\caption{Three adaptation mechanisms in AIC-VDS}
\label{tab:adaptation_mechanisms}
\centering
\scriptsize
\setlength{\tabcolsep}{2pt}

\begin{tabular}{lccc}
\toprule
Mechanism & Train? & Method & Applies to \\
\midrule
LLM fine-tuning & No & -- & Not used \\
In-context learning & No & Prompt & Frozen LLM \\
Attention/velocity & Yes & Value regression & Scoring modules \\
\bottomrule
\end{tabular}

\end{table}

The per-UAV, per-decision-epoch inference complexity of the proposed AIC-VDS is
\begin{equation}
\label{250}
\mathcal{O}\!\left(Jd_xd+J^2d\right)
+
\mathcal{O}\!\left(
n_{\mathrm{LLM}}
\left[L^2d_{\mathrm{LLM}}+Ld_{\mathrm{LLM}}^2\right]
\right),
\end{equation}
where
\begin{itemize}
    \item $\mathcal{O}(Jd_xd+J^2d)$ corresponds to the attention module operating over $J$ sensors, where $d_x$ and $d$ denote the input-feature and attention hidden dimensions, respectively;
    \item $\mathcal{O}\!\left(n_{\mathrm{LLM}}[L^2d_{\mathrm{LLM}}+Ld_{\mathrm{LLM}}^2]\right)$ represents the generic Transformer inference complexity, where $L$ is the compressed-prompt length, $d_{\mathrm{LLM}}$ is the internal hidden dimension, and $n_{\mathrm{LLM}}$ is the number of Transformer layers.
\end{itemize}
Thus, assuming sequential execution, the aggregate per-decision-epoch complexity across $I$ UAVs is
\begin{equation}
\label{251}
\mathcal{O}\!\left(
I\left[
Jd_xd+J^2d+
n_{\mathrm{LLM}}
\left(L^2d_{\mathrm{LLM}}+Ld_{\mathrm{LLM}}^2\right)
\right]
\right).
\end{equation}

\begin{table*}[!t]
\centering
\caption{Simulation, Environment, and AIC-VDS Hyperparameter Settings}
\label{tab:general_params}
\renewcommand{\arraystretch}{1.1}
\begin{tabular}{|p{7cm}|p{2.5cm}|}
\hline
\textbf{Parameter} & \textbf{Value} \\
\hline
\multicolumn{2}{|l|}{\textit{Shared Simulation \& Environment Settings}} \\
\hline
Number of independent runs / seeds & 10 \\
\hline
Timesteps per run (rollout length) & 30 \\
\hline
Number of UAVs & 3 \\
\hline
Number of ground sensors (\texttt{num\_gr}) & 10 \\
\hline
Monitoring area (areax $\times$ areay) & $1000 \times 1000$ \\
\hline
Initial sensor energy (\texttt{IE}) & 50 J \\
\hline
Max queue length (\texttt{maxqlen}) & 60 packets \\
\hline
Number of velocity levels & 15 (0--14) \\
\hline
Default / base UAV altitude & 100 \\
\hline
UAV radius (\texttt{radius}) & 360 \\
\hline
\multicolumn{2}{|l|}{\textit{Channel / Path-Loss Model (Shared)}} \\
\hline
$\eta_{LoS}$ & 1 \\
\hline
$\eta_{NLoS}$ & 20 \\
\hline
$a$ (LoS model constant) & 20 \\
\hline
$\kappa$ (LoS model constant) & 0.3 \\
\hline
Carrier frequency (\texttt{fc}) & 2000 \\
\hline
Speed of light (\texttt{vc}) & $3 \times 10^{8}$ \\
\hline
\multicolumn{2}{|l|}{\textit{AIC-VDS: Sensor Urgency Weighting}} \\
\hline
$\alpha$ (\texttt{urgency\_alpha}) & 0.5 \\
\hline
$\beta$ (\texttt{urgency\_beta}) & 0.5 \\
\hline
\multicolumn{2}{|l|}{\textit{AIC-VDS: Attention / Velocity Modules}} \\
\hline
Attention input dimension & 5 \\
\hline
Attention hidden dimension & 32 \\
\hline
Optimizer & Adam \\
\hline
Learning rate & $5 \times 10^{-4}$ \\
\hline
Top-$k$ sensors retained & $k = 5$ \\
\hline
\multicolumn{2}{|l|}{\textit{AIC-VDS: LLM Query}} \\
\hline
Model & GPT-4o-mini \\
\hline
Examples per query (\texttt{num\_examples}) & 30 \\
\hline
\end{tabular}
\end{table*}

\begin{table*}[!t]
\centering
\caption{Hyperparameter Settings for the DRL Baselines (MADQN and MAPPO)}
\label{tab:baseline_params}
\renewcommand{\arraystretch}{1.1}
\begin{tabular}{|p{7cm}|p{2.5cm}|}
\hline
\textbf{Parameter} & \textbf{Value} \\
\hline
\multicolumn{2}{|l|}{\textit{Shared DRL Baseline Settings (MADQN \& MAPPO)}} \\
\hline
Discount factor ($\mu$) & 0.99 \\
\hline
Optimizer & Adam \\
\hline
Network architecture (shared trunk) & FC (400 $\rightarrow$ 300) \\
\hline
Action distribution & Categorical (10-node, 15-velocity) \\
\hline
\multicolumn{2}{|l|}{\textit{MADQN-Specific Settings}} \\
\hline
Max. episodes & 1000 \\
\hline
Target network soft-update rate ($\tau$) & 0.005 \\
\hline
Target network update frequency & every 3 iterations \\
\hline
Learning rate & $5 \times 10^{-4}$ \\
\hline
Batch size & 100 \\
\hline
Replay buffer max. size & $1 \times 10^{6}$ \\
\hline
Q-network output heads & 2 \\
\hline
\multicolumn{2}{|l|}{\textit{MAPPO-Specific Settings}} \\
\hline
Learning rate & $3 \times 10^{-4}$ \\
\hline
PPO clip range ($\epsilon$) & 0.2 \\
\hline
Entropy coefficient & 0.01 \\
\hline
Value loss coefficient & 0.5 \\
\hline
PPO epochs per update & 4 \\
\hline
\end{tabular}
\end{table*}
\begin{figure*}[t]
\centering
\begin{subfigure}[t]{0.32\textwidth}
    \centering
    \includegraphics[width=\linewidth, height=4.5cm, keepaspectratio]{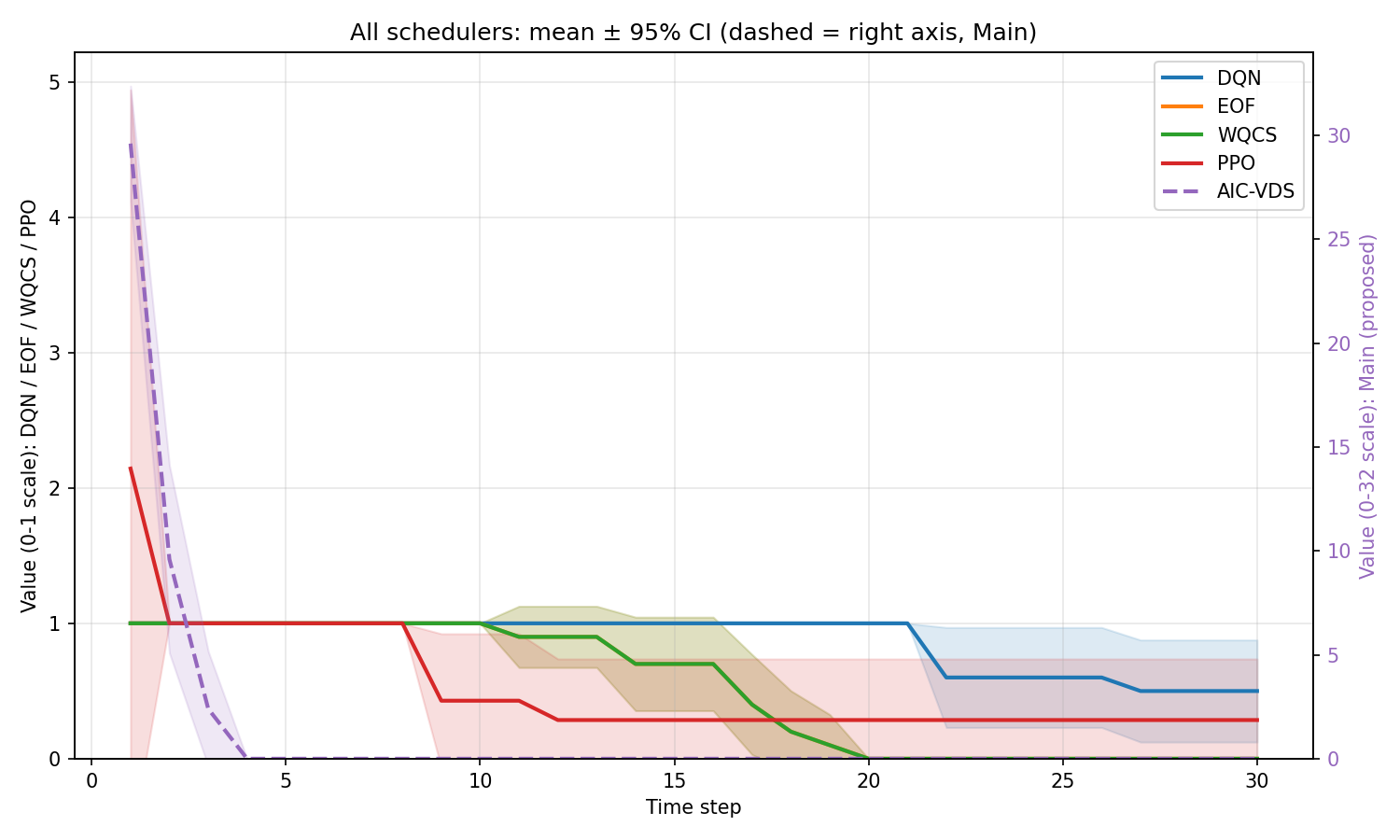}
    \caption{}
    \label{fig:scheduler}
\end{subfigure}\hfill
\begin{subfigure}[t]{0.32\textwidth}
    \centering
    \includegraphics[width=\linewidth, height=4.5cm, keepaspectratio]{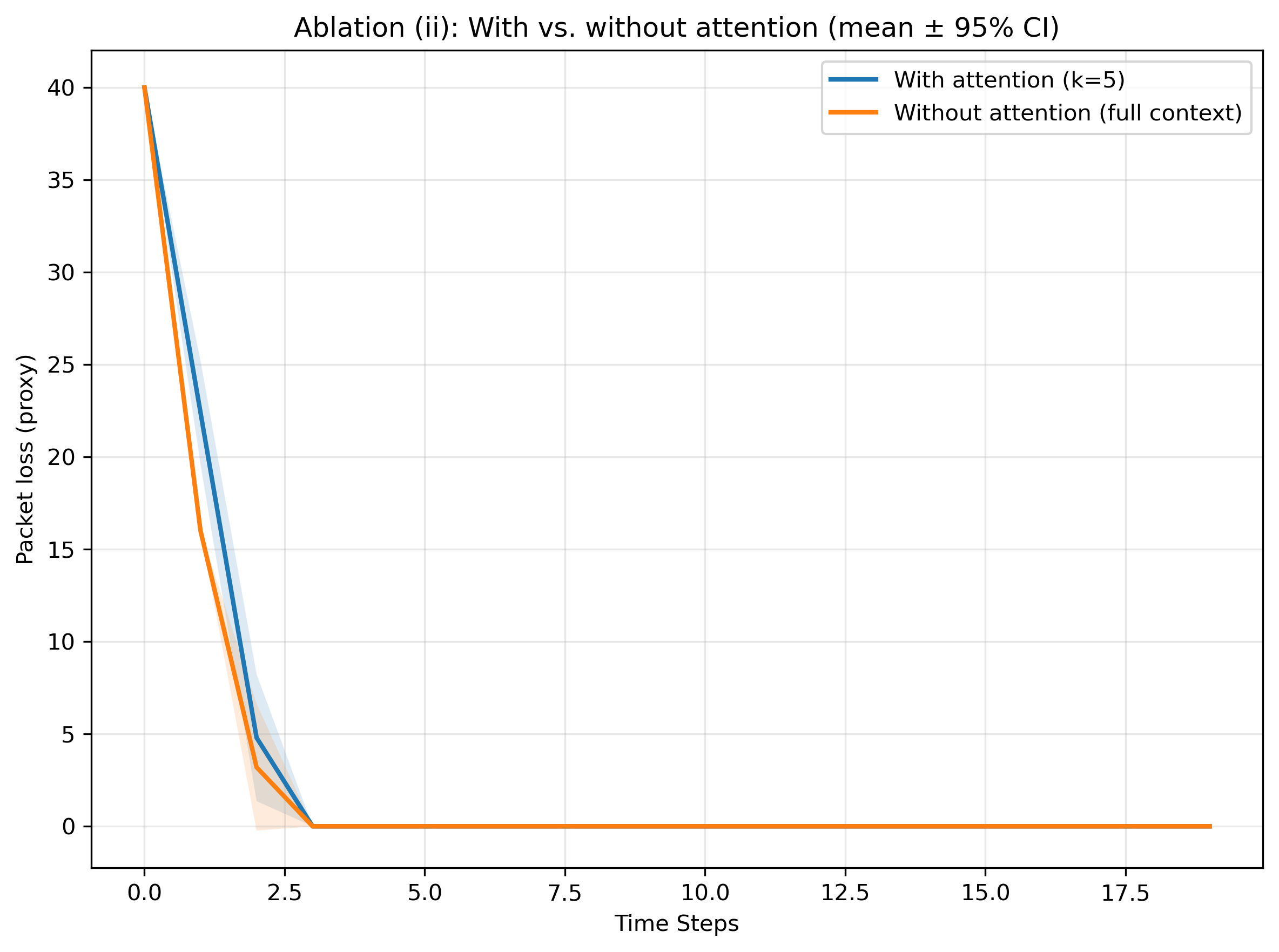}
    \caption{}
    \label{fig:ablation-attention}
\end{subfigure}\hfill
\begin{subfigure}[t]{0.32\textwidth}
    \centering
    \includegraphics[width=\linewidth, height=4.5cm, keepaspectratio]{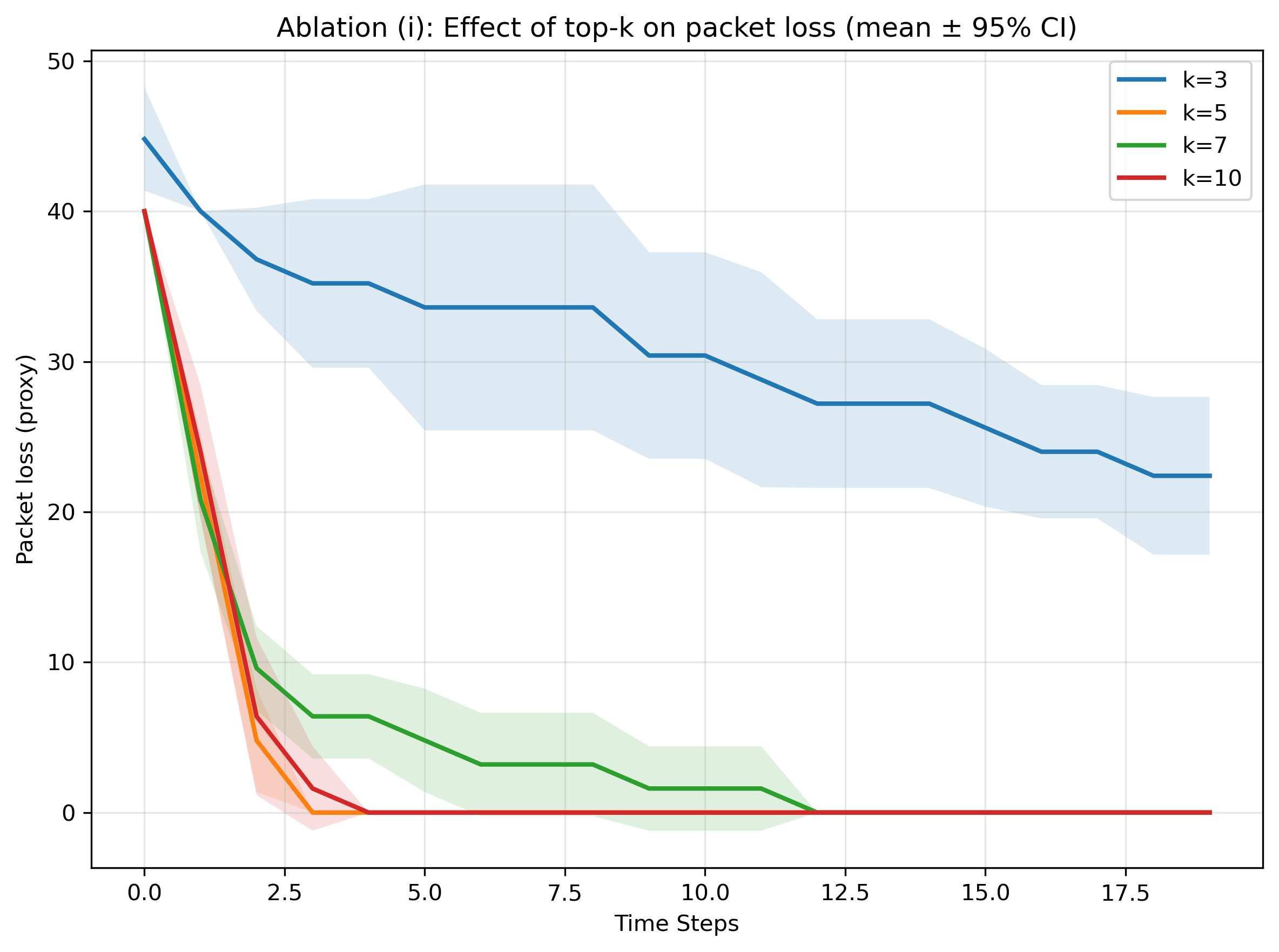}
    \caption{}
    \label{fig:ablation-topk}
\end{subfigure}
\caption{(a) Performance trajectories for AIC-VDS and the MADQN, EOF, WQCS, and MAPPO baselines;  (b) Attention on/off comparison.(c) Effect of top-$k$ over time;}
\label{fig:ablation-all}
\end{figure*}
\begin{figure*}[htbp]
    \centering
    \includegraphics[width=0.8\textwidth]{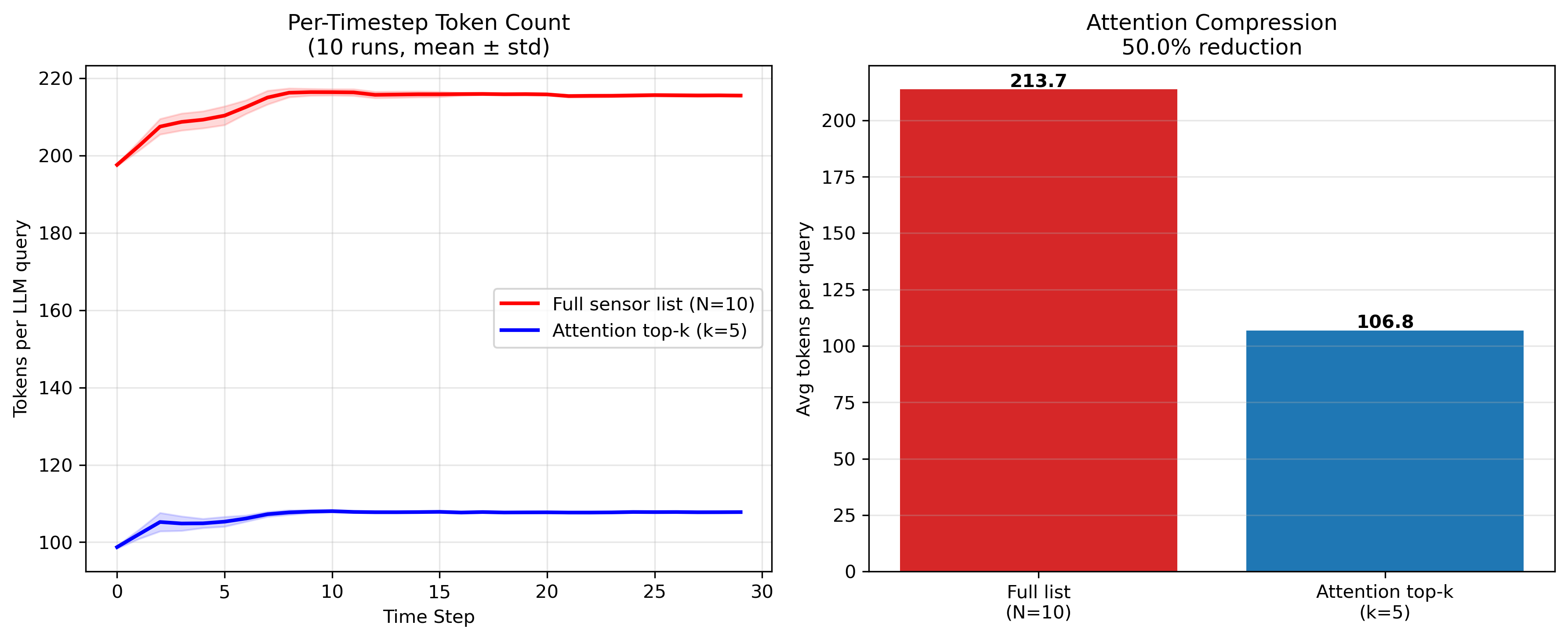}
\caption{Prompt-token reduction achieved by the proposed attention-based compression module. (Left) Token count per LLM query at each time step of the 30-step simulation horizon, averaged over 10 independent runs, with shaded regions denoting $\pm 1$ standard deviation. The uncompressed full sensor list is compared with the top-$k$ ($k=5$) sensor representation selected by the attention module. (Right) Average tokens per query across the full horizon for each configuration, showing a reduction from 213.7 to 106.8 tokens, corresponding to an approximately 50\% decrease in prompt length.}
    \label{fig:myfigure20}
\end{figure*}
\begin{figure*}[htbp]
    \centering

    \begin{subfigure}{0.32\textwidth}
        \centering
        \includegraphics[width=\linewidth]{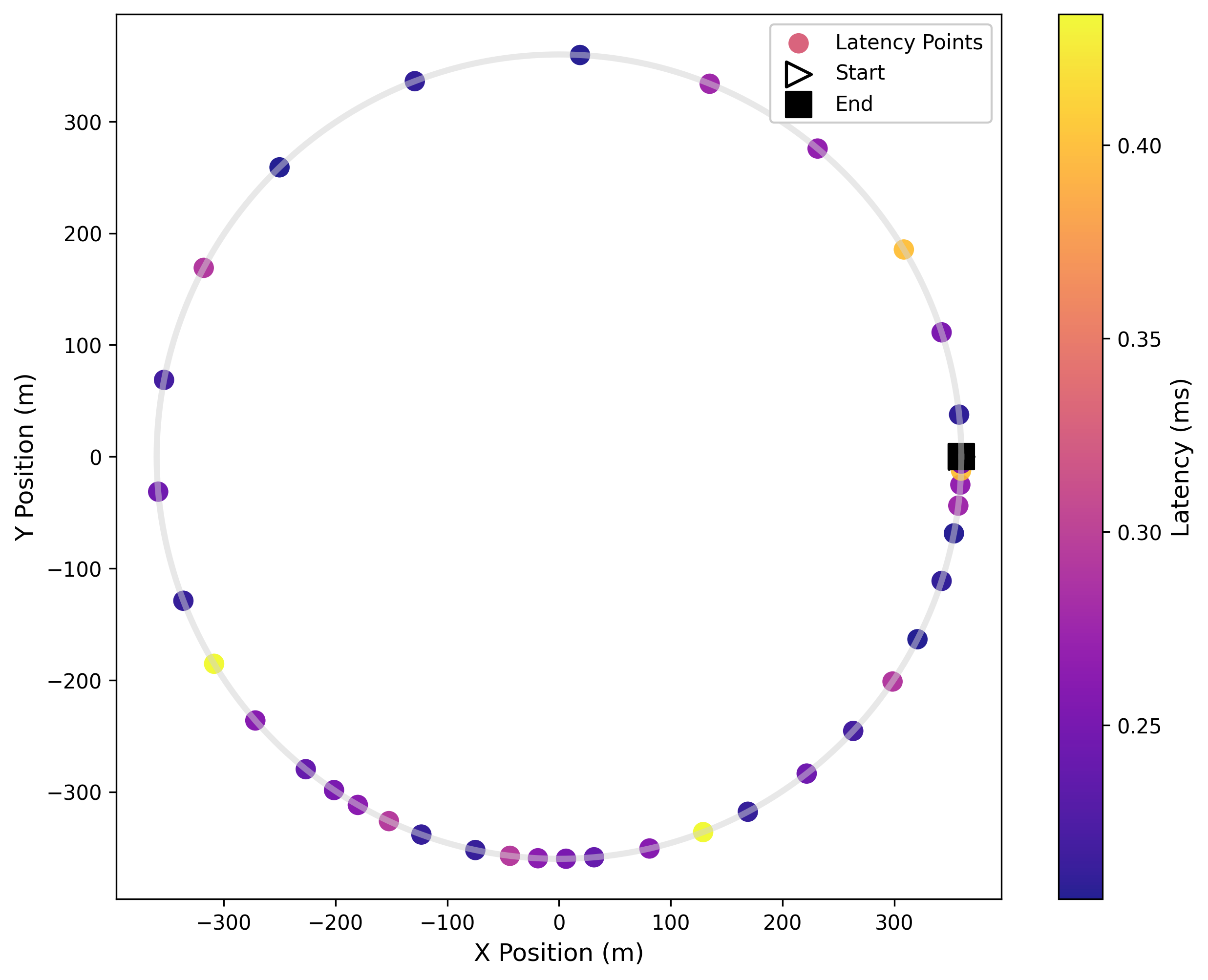}
        \caption{}
        \label{fig:fig1}
    \end{subfigure}
    \hfill
    \begin{subfigure}{0.32\textwidth}
        \centering
        \includegraphics[width=\linewidth]{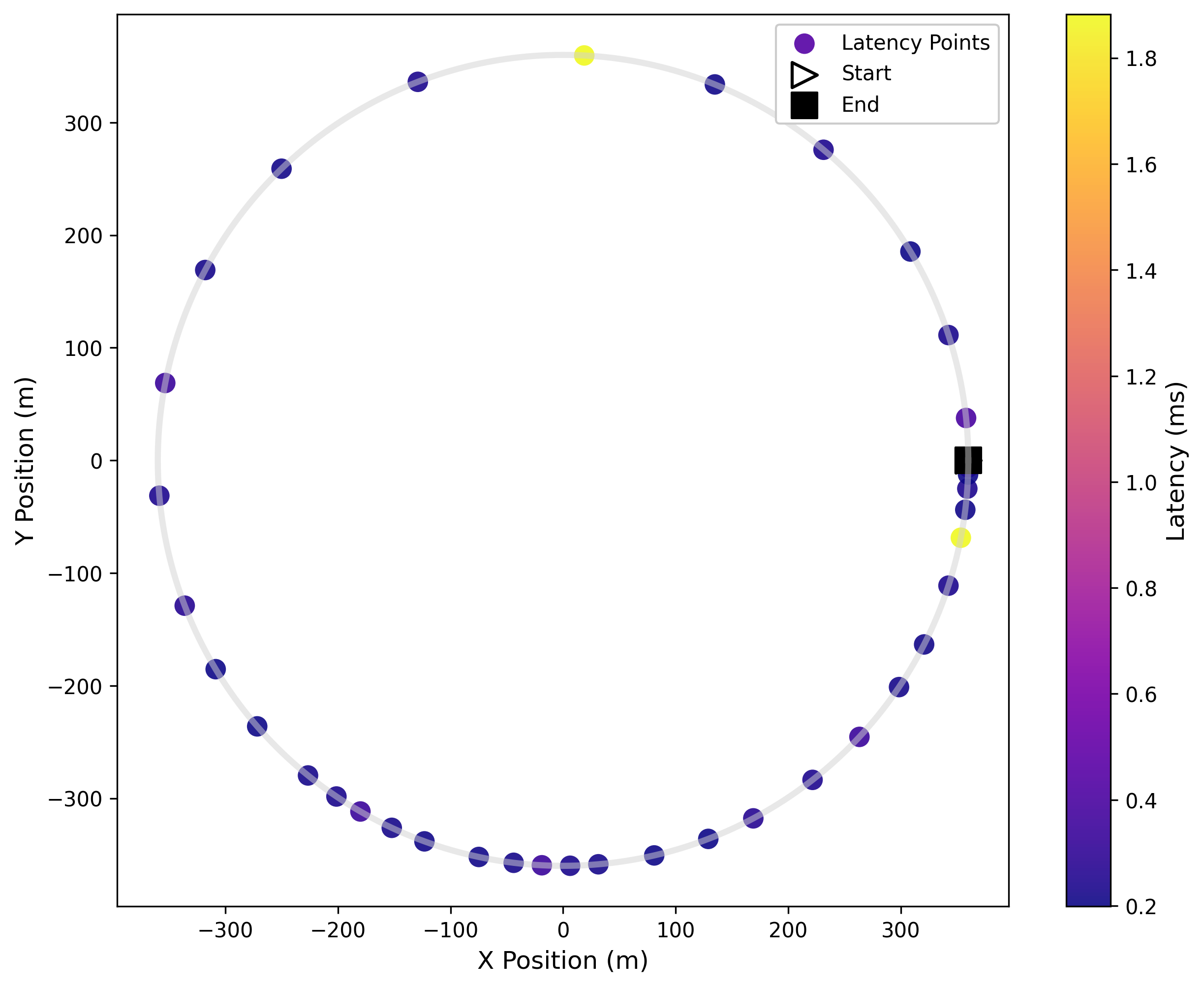}
        \caption{}
        \label{fig:fig2}
    \end{subfigure}
    \hfill
    \begin{subfigure}{0.32\textwidth}
        \centering
        \includegraphics[width=\linewidth]{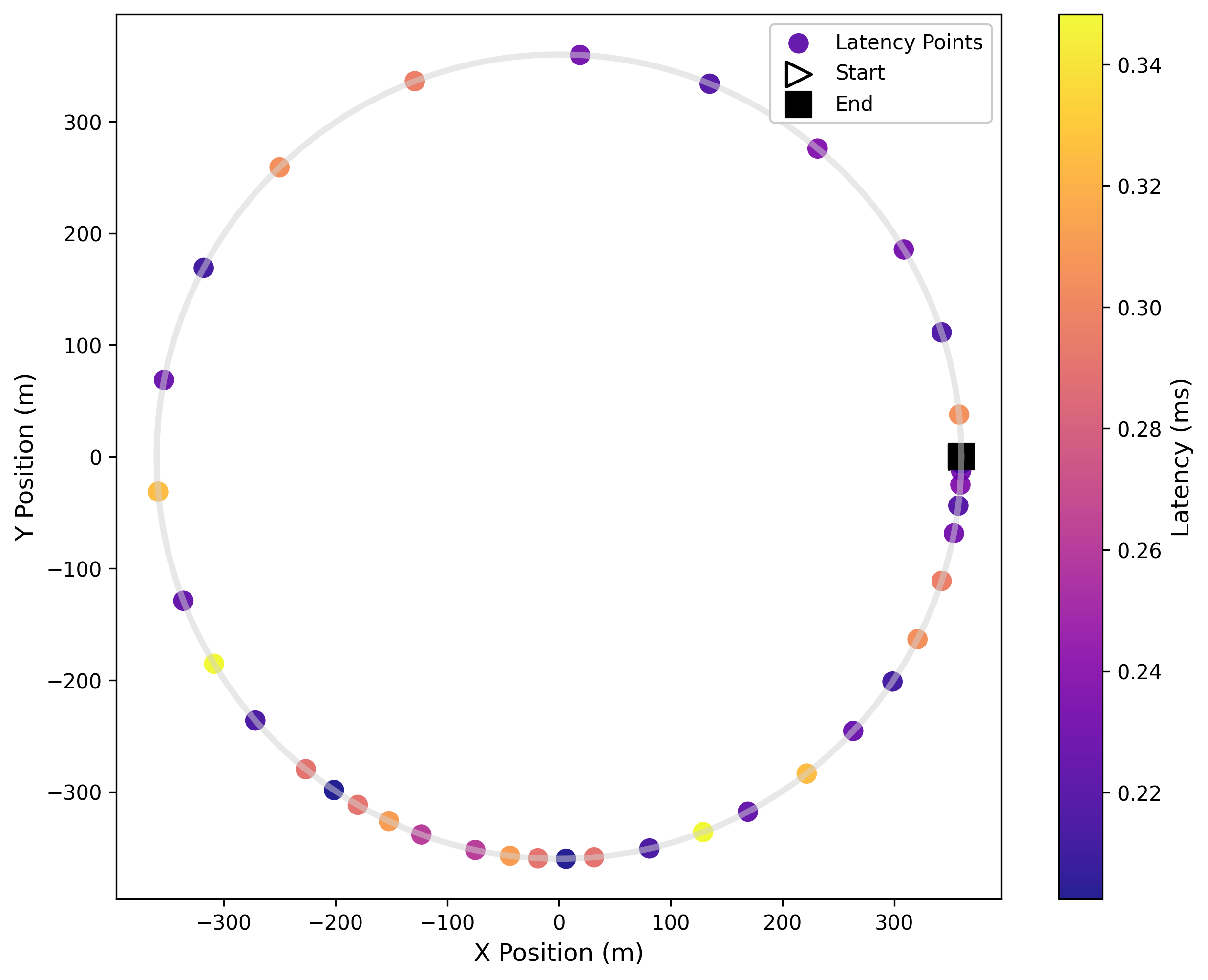}
        \caption{}
        \label{fig:fig3}
    \end{subfigure}

\caption{Per-query round-trip LLM latency along the circular flight trajectory for each of the three UAVs. Each marker denotes a decision epoch along the trajectory, color-coded by the round-trip latency (ms) measured at that epoch, with the start and end waypoints indicated by triangle and square markers, respectively. The latency color scale differs across subfigures to preserve visual resolution; therefore, colors should not be compared directly across UAVs.}
    \label{fig:three_figures}
\end{figure*}

\section{Numerical Results and Discussion}
\label{sec6}
This section presents the simulation setup and evaluates the performance of the proposed AIC-VDS framework for multi-UAV-assisted pipeline monitoring.

\subsection{Implementation of AIC-VDS}
The main simulation parameters are summarized in Tables~\ref{tab:general_params} and \ref{tab:baseline_params}. We consider a pipeline-monitoring scenario with 10 sensors randomly deployed over a $1000~\text{m} \times 1000~\text{m}$ area. Each sensor has a battery capacity of $50~\text{J}$, a data-queue capacity of 60 packets, and a maximum transmission power of $100~\text{mW}$. Each simulation episode comprises 30 time steps.
\par
The proposed AIC-VDS framework is implemented in a custom Python simulation environment. UAVs collect monitoring data from distributed pipeline sensors and adaptively select sensors and determine their flight velocities using the attention module and the frozen-LLM ICL mechanism. The UAV--sensor wireless links are modeled according to the probabilistic channel model described in Section~\ref{sec4}. All experiments are conducted on a Lenovo workstation running Ubuntu 20.04 LTS, equipped with an Intel Core i5-7200U CPU operating at $2.50~\text{GHz}$ and $16~\text{GB}$ of RAM.

\subsection{Baselines Descriptions}

\begin{enumerate}
\item \textbf{MADQN\cite{emami2021joint}:}
MADQN is selected as a DRL-based benchmark for data-collection scheduling and velocity control. Each UAV is controlled by an independent agent that learns a policy to minimize packet loss based on key network-state variables, including the UAV position, sensor queue lengths, battery levels, and channel conditions. Three UAVs follow predefined circular trajectories to collect data from 10 pipeline-monitoring sensors. The DQN-based agents jointly determine sensor-scheduling and velocity decisions using a two-branch network architecture.

\item \textbf{MAPPO}:
MAPPO is selected as a DRL-based benchmark for data-collection scheduling and velocity control. Each UAV executes a decentralized policy trained to minimize packet loss based on network-state variables, including UAV positions, sensor queue lengths, battery levels, and channel conditions. Three UAVs follow predefined circular trajectories to collect data from 10 pipeline-monitoring sensors.
\item 
\textbf{EOF:}
The EOF scheduler is a heuristic baseline that estimates each sensor's time-to-overflow from observed queue growth over successive decision epochs, then combines this urgency with channel quality to compute a priority score. At each time step, each UAV selects the highest-priority sensor and sets its velocity based on its distance to that sensor.

\item 
\textbf{WQCS:}
WQCS is a heuristic baseline that ranks sensors using a priority score equal to the product of queue length and achievable transmission rate, thereby favoring sensors with larger backlogs and better channel conditions. At each decision epoch, each UAV independently selects the highest-priority sensor and adjusts its velocity along its predefined trajectory to serve the selected sensor.
\end{enumerate}

\subsection{Performance Evaluation}

Fig.~\ref{fig:scheduler}
compares the performance trajectories of AIC-VDS and four baseline schedulers – MADQN, MAPPO, EOF, and WQCS – over a 30-step simulation horizon. The MADQN and MAPPO curves correspond to their performance during the final training episode. AIC-VDS achieves zero packet loss by time step 4, five to seven time steps earlier than EOF/WQCS (time step 20) and MADQN (time step 27), while MAPPO maintains relatively stable packet-loss performance. This faster stabilization comes at the cost of a larger initial transient loss, which the baselines do not exhibit. These results suggest that the main advantage of AIC-VDS is its faster stabilization after the initial transient and minimal packet loss.
\par
Fig.\ref{fig:ablation-attention}
compares data collection schedules with and without the learned attention mechanism, reporting mean packet loss across 5 runs (20 timesteps) with 95\% confidence intervals. Over the initial transient (timesteps 0–3), both configurations show comparable performance with similar patterns. In particular, the configuration without attention tracks marginally below the attention-enabled curve, with overlapping confidence intervals throughout, indicating no statistically meaningful difference in convergence speed or final performance.
\par
Fig.~\ref{fig:ablation-topk} shows that the choice of the top-$k$ sensor selection parameter directly affects the convergence performance of the AIC-VDS data collection schedule. A sufficiently large sensor context ($k \geq 5$) enables rapid reduction of packet loss and stable convergence. In contrast, an overly restrictive selection ($k = 3$) limits the information available to the LLM and leads to slower, less consistent optimization. Since increasing $k$ beyond 5 provides negligible improvement, $k = 5$ achieves a favorable trade-off between sensor information availability and input complexity.
\par
Fig.~\ref{fig:myfigure20} confirms that the proposed attention-based sensor selection method achieves substantial LLM prompt compression by retaining only the most informative sensor states. The top-$k$ ($k=5$) representation reduces the input length by 50.0\% while maintaining stable query behavior, demonstrating an effective trade-off between information preservation and LLM communication and processing overhead.
\par
Fig.~\ref{fig:three_figures} shows that LLM inference latency remains generally low and spatially consistent throughout the UAV trajectories, with only a few isolated excursions. Although individual UAVs experience different latency ranges due to occasional transient spikes, no persistent spatial correlation between UAV position and LLM latency is observed. This indicates that the proposed LLM-based decision-making framework maintains reliable query responsiveness during UAV data collection, with latency variations primarily driven by intermittent processing or communication fluctuations rather than trajectory-dependent effects.

\begin{figure}[htbp]
    \centering
    \includegraphics[width=9cm,height=8cm]{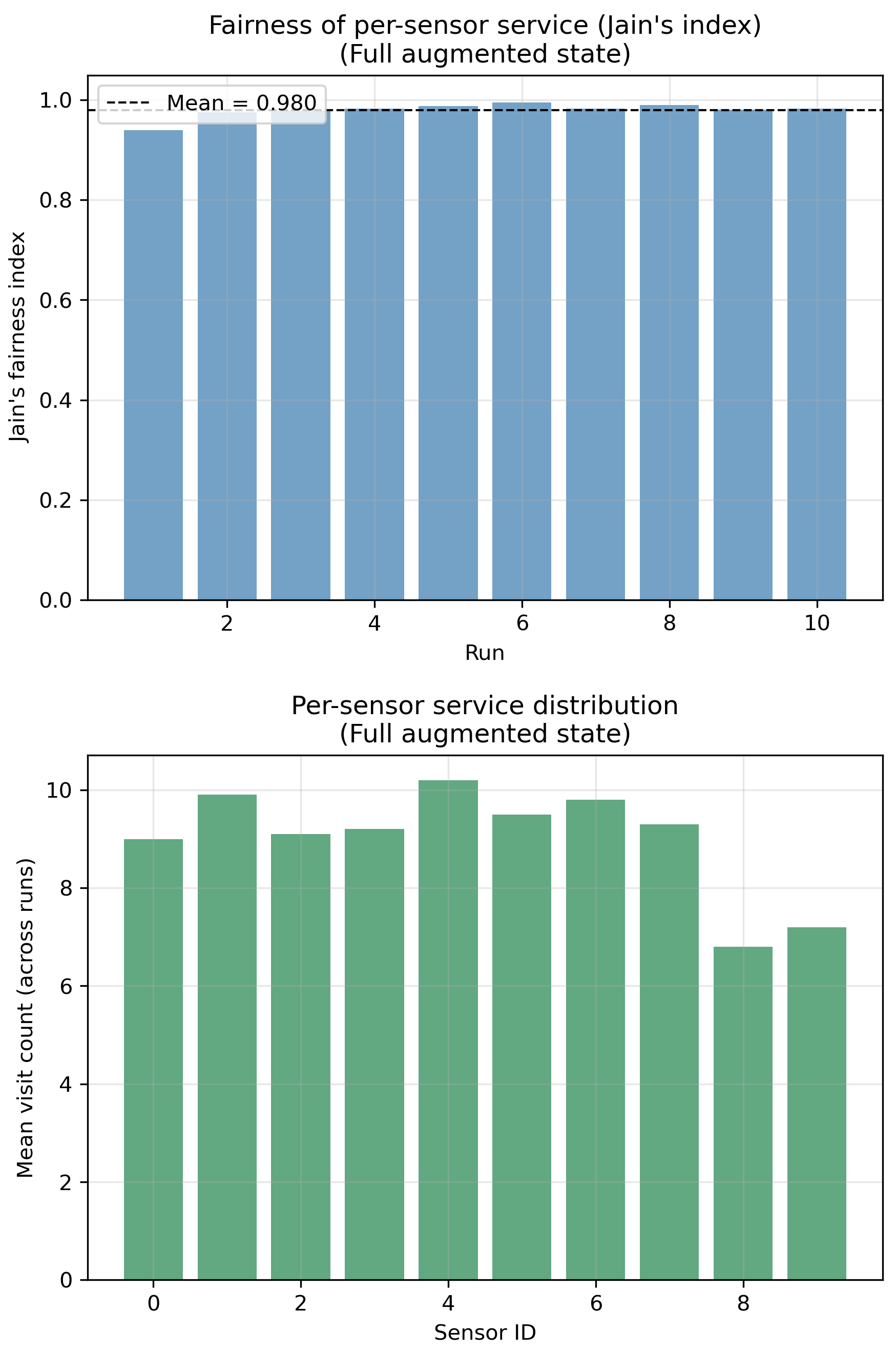}
\caption{Fairness performance of the proposed full augmented-state approach over ten independent simulation runs. (up) Jain's Fairness Index (JFI) for each run, showing consistently high fairness with a mean value of 0.980, indicating near-uniform service allocation among all sensors. (down) Average number of visits received by each sensor across all runs, demonstrating a balanced service distribution with only minor variations among sensors. The results confirm that the proposed method achieves equitable and stable resource allocation while avoiding significant long-term service bias.}
   \label{fig:myfigure167}
\end{figure}
Fig.~\ref{fig:myfigure167} shows that the proposed full augmented-state representation enables fair and balanced sensor service allocation. The near-optimal Jain’s Fairness Index (0.980) and the consistent per-sensor visit distribution across independent runs demonstrate that the learned policy effectively avoids sensor starvation while maintaining an equitable workload distribution. The small variations among individual sensors have a negligible impact on overall fairness, confirming the robustness of the proposed approach in achieving long-term balanced data collection.

\section{Conclusion} \label{sec7}
In this paper, we proposed the AIC-VDS framework to address the joint scheduling and mobility challenges in multi-UAV-assisted pipeline monitoring.
To mitigate the substantial input overhead of LLMs, AIC-VDC utilizes an attention-based context pruning mechanism.
This module preserves critical, task-relevant sensory information while significantly reducing input redundancy and prompt lengths. By integrating this compressed state representation with ICL on a frozen edge-hosted LLM, the framework dynamically generates adaptive velocity-control and data-collection schedules without requiring computationally expensive model fine-tuning. 
Simulation results demonstrate that, following a rapid initial convergence, AIC-VDS consistently outperforms the MADQN, MAPPO, EOF, and WQCS baselines in minimizing packet loss.

\bibliographystyle{IEEEtran}
\bibliography{references}
\end{document}